\documentclass[runningheads]{llncs}

\usepackage{eccv}

\usepackage{eccvabbrv}

\usepackage{graphicx}
\usepackage{booktabs}
\usepackage{algorithm}
\usepackage{algpseudocode}
\usepackage{comment}
\usepackage{mathtools}
\usepackage{cancel}

\usepackage{arydshln}
\usepackage{wrapfig}

\usepackage[accsupp]{axessibility}  

\usepackage{hyperref}

\usepackage{orcidlink}
\usepackage{multirow}
\usepackage{ntheorem}
\newcommand{\blockcomment}[1]{}
\newtheorem*{theorem*}{Theorem}

\definecolor{darkspringgreen}{rgb}{0.09, 0.45, 0.27}

\newcommand{\SK}[1]{\textcolor{red}{[SK: #1]}}

\newcommand{\FY}[1]{\textcolor{green}{[FY: #1]}}

\usepackage{titletoc}
\usepackage[section]{placeins}
\usepackage{twoopt}
\usepackage{tikz}
\usetikzlibrary{fit}
\newcommand\tikzmark[1]{%
  \tikz[remember picture,overlay]\node[inner xsep=0pt] (#1) {};}
\newcommandtwoopt\Textbox[6][3.5cm][4.7cm]{%
\begin{tikzpicture}[remember picture,overlay]
  \coordinate (aux) at ([xshift=#1]#4);
  \node[inner ysep=3pt,yshift=0.6ex,
    fit=(#3) (aux),baseline] 
    (box) {};
  \node[text width=#2,anchor=north east,
    font=\sffamily\scriptsize,align=right] 
    at (box.north east) {#5};
\end{tikzpicture}%
}
\begin{document}

\title{Bucketed Ranking-based Losses for \\
Efficient Training of Object Detectors} 

\titlerunning{Bucketed Ranking-based Losses}

\author{Feyza Yavuz\inst{1}\orcidlink{0000-1111-2222-3333} \and
Baris Can Cam\inst{1}\orcidlink{1111-2222-3333-4444} \and
Adnan Harun Dogan\inst{1}\orcidlink{0000-0002-6802-6405} \and
Kemal Oksuz\inst{2}\orcidlink{0000−0002−0066−1517} 
\and
Emre Akbas\inst{1,3,*}\orcidlink{0000-0002-3760-6722} \and
Sinan Kalkan\inst{1,3,*}\orcidlink{0000-0003-0915-5917}}

\authorrunning{Yavuz et al.}

\institute{Department of Computer Engineering, METU, Ankara, Türkiye \and 
Five AI Ltd., United Kingdom \and
METU ROMER Robotics Center, Ankara, Türkiye\\
* Equal senior authorship}
\maketitle
\begin{abstract}

Ranking-based loss functions, such as Average Precision Loss and Rank\&Sort Loss, outperform widely used score-based losses in object detection. These loss functions better align with the evaluation criteria, have fewer hyperparameters,  
and offer robustness against the imbalance between positive and negative classes. However, they require pairwise comparisons among $P$ positive and $N$ negative predictions,  introducing a  time complexity of $\mathcal{O}(PN)$, which is prohibitive since $N$ is often large (e.g., \(10^8\) in ATSS). Despite their advantages, the widespread adoption of ranking-based losses has been hindered by their high time and space complexities.
In this paper, we focus on improving the efficiency of ranking-based loss functions. To this end, we propose Bucketed Ranking-based Losses which group negative predictions into $B$ buckets ($B \ll N$) in order to reduce the number of pairwise comparisons so that time complexity can be reduced. Our method enhances the time complexity, reducing it to $\mathcal{O}(\max (N \log(N), P^2))$. To validate our method and show its generality, we conducted experiments on 2 different tasks, 3 different datasets, 7 different detectors.  
We show that Bucketed Ranking-based (BR) Losses yield the same accuracy with the unbucketed versions and provide $2\times$ faster training on average. 
We also train, for the first time, transformer-based object detectors using ranking-based losses, thanks to the efficiency of our BR. When we train CoDETR, a state-of-the-art transformer-based object detector, using our BR Loss, we consistently outperform its original results over several different backbones. 
Code is available at: \url{https://github.com/blisgard/BucketedRankingBasedLosses}. 
\end{abstract} 
\section{Introduction}
\label{sec:intro}



    

\begin{figure}[hbt!]
    
    \includegraphics[width=1.0\columnwidth]{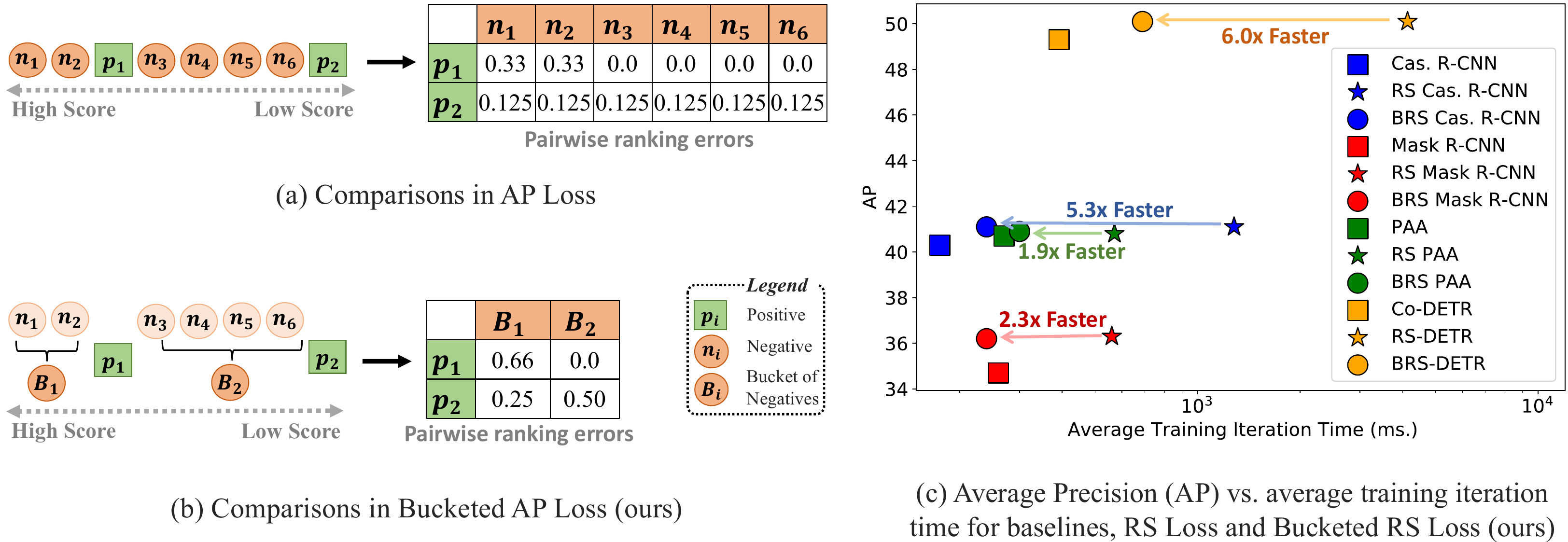}
    \caption{(a) Existing ranking-based losses (i.e., AP Loss \cite{APLoss1,APLoss2}) incur significant overhead owing to pairwise comparisons between positives and negatives. (b) We propose bucketing negatives to decrease the number of comparisons, and hence the complexity. Under certain assumptions, our bucketing approach provides equal gradients with conventional ranking-based losses such as AP Loss in (a). (c) Accuracy and efficiency comparison. Our BRS formulation facilitates faster (between $1.9 \times$ and $6.0 \times$) training of visual detectors with similar AP.}
    \label{fig:teaser}
\end{figure}

Training an object detector incurs many challenges unique to object detection, including
(i) assigning dynamically a large number of object hypotheses with the objects  \cite{ATSS,paa,sparsercnn,FasterRCNN,CascadeRCNN,CornerNet,maIoU}, (ii) sampling among these hypotheses to ensure that the background class does not dominate training \cite{SSD, FasterRCNN,OHEM,LibraRCNN} and (iii) minimizing a multi-task objective function \cite{FasterRCNN,aLRPLoss,RSLoss,FocalLoss}.
While the choices on assignment and sampling generally vary depending on the trained object detector, it is very common to combine Cross Entropy or Focal Loss \cite{FocalLoss} with a regression loss \cite{GIoULoss,DIoULoss} as the multi-task loss function.
Recently proposed ranking-based loss functions \cite{RSLoss,aLRPLoss,APLoss1,APLoss2} offer an alternative approach for addressing these challenges by formulating the training objective based on the rank of positive examples over negative examples. 

\noindent\textbf{Benefits of ranking-based losses.} 
First, they are inherently robust to imbalance \cite{aLRPLoss} and hence, do not require any sampling mechanism under very challenging scenarios \cite{RSLoss}, e.g., even when the background-foreground ratio is 10K for LVIS \cite{LVIS}.
Second, they are shown to generalize over different detectors with diverse architectures -- with the exception of transformer-based ones \cite{DETR,DeformableDETR,CoDETR}, since ranking-based losses further slow down 
the training of transformer-based detectors, which we address in this paper. 
Such losses also offer significant performance gain over their score-based counterparts, and having less hyperparameters, they are much easier to tune \cite{aLRPLoss,RSLoss}. 
%

\noindent\textbf{The drawback of ranking-based losses.} Compared to score-based losses, ranking-based losses are more inefficient as the ranking operation inherently requires each pair of object hypotheses to be compared against each other (Fig. \ref{fig:teaser}(a)), inducing a quadratic complexity on the large number of object hypotheses (e.g., $10^{8}$ for ATSS \cite{ATSS} on COCO  \cite{COCO}).
As a result, vectorized implementations for parallel GPU processing are infeasible as such large matrices (e.g., with $\sim10^{16}$ entries for ATSS) do not fit into GPU memories.
This has driven researchers towards alternative ways of computing these matrices, which, in the end, saves from the storage complexity but results in more inefficient algorithms \cite{APLoss1,APLoss2}.

\noindent\textbf{Our approach.} The main cause for the aforementioned complexity is the background class, i.e., negatives, forming up to $99.9\%$ of the hypotheses \cite{oksuz2020imbalance}.
Considering this, we introduce a novel bucketing approach (Fig. \ref{fig:teaser}(b)) on these negatives to address the main shortcoming of the ranking-based loss functions.
Specifically, we first sort all examples based on their scores and bucket the negatives located between successive positives. 
The negatives within the same bucket are treated as a single example and represented by their average score. 
We then incorporate our bucketing approach into AP Loss \cite{APLoss1,APLoss2} and RS Loss \cite{RSLoss}, and introduce novel mechanisms for calculating the gradients correctly.

\noindent\textbf{The significance of our approach.} Theoretically, we prove that our bucketing approach provides the same gradients as the original ranking-based loss. Practically, we show we can reduce loss computation time by up to $\sim 40\times$ and training time of a detector by up to $6\times$ (Fig. \ref{fig:teaser}(c)), addressing the main drawback of ranking-based loss functions.
%
%
%
Hence, we show that the gap between the score-based and ranking-based loss functions in terms of training time vanishes, enabling us to train strong transformer-based object detectors such as Co-DETR using our loss functions.
As ranking-based loss functions now surpass score-based ones or perform on par in every aspect such as accuracy, tuning simplicity, robustness to imbalance and training time, we believe that our work will make their prevalence more dominant in the upcoming object detectors.

\noindent\textbf{Contributions}. Our contributions can be summarized as follows: 
\begin{itemize}
    \item We propose a novel bucketing approach to improve the efficiency of computa\-tionally-expensive ranking-based losses to train object detectors. Theoretically, our approach yields  the same gradients with the original ranking-based losses while decreasing their time complexity. Practically, we enable up to $6 \times$ faster training using our bucketing approach with no accuracy loss. 
    \item For the first time, we incorporate ranking-based loss functions to transformer-based detectors. Specifically, we construct BRS-DETR by replacing the training objective of the state-of-the-art transfomer-based object detector Co-DETR \cite{CoDETR} by our Bucketed RS (BRS) Loss. 
    \item Our comprehensive experiments on detection and instance segmentation on 3 different challenging datasets, 5 backbones and 7 detectors show the effectiveness and generalizability of our approach. Our BRS-DETR reaches $50.1$ AP on COCO val set with only 12 epochs and ResNet-50 backbone, outperforming all existing detectors using the same backbone and 300 queries. BRS-DETR also outperforms CoDETR with Swin-T and Swin-L backbones, reaching $57.2$ AP on the COCO dataset. 
\end{itemize}

\section{Related Work}
\label{sec:related_work}

Although score-based loss functions, such as cross-entropy and focal loss, are widely used to train both CNN-based \cite{CascadeRCNN, FocalLoss, FreeAnchor} and transformer-based object detectors \cite{DETR, DeformableDETR, CoDETR}, ranking-based losses for visual detection offer advantages. The primary advantage is their robustness to class imbalance \cite{aLRPLoss}. They yield on par or better performance compared to score-based losses, without the need for extensive tuning of class weights or multistage sampling techniques. Another advantage they offer is the ease of balancing losses among different tasks. In AP Loss \cite{APLoss1,APLoss2}, aLRP Loss \cite{aLRPLoss} and RS Loss \cite{RSLoss}, the authors use the ratio of classification and regression losses to balance these tasks. As a result, ranking-based losses have much fewer hyperparameters during training, compared to score-based ones. 

\noindent\textbf{Other Ranking-based Losses.} The literature has explored other ranking-based loss functions with different goals and assumptions. For example, DR Loss \cite{DRLoss} constrains a margin to be satisfied between positives and negatives, not taking the recall into account. Smooth AP \cite{smoothAP} introduces a smooth, differentiable approximation of AP, with the assumption that $|\mathcal{N}|$ is not large. Several studies focused on improving the efficiency of ranking-based losses for use in Support Vector Machines (SVMs) \cite{OptimizingUpperBound,mohapatra2014efficient}. However, these solutions have been limited to training linear SVMs and hence, problems that can be solved using linear functions. In this paper, we focus on improving ranking-based loss functions \cite{APLoss1,APLoss2,RSLoss} that can be used for training complex deep networks such as object detectors.
 
\noindent\textbf{Comparative Summary.}
Ranking-based training objective for visual detectors proved their superiority over score-based loss functions in visual detection tasks \cite{APLoss1,APLoss2, aLRPLoss, RSLoss}. 
However, due to the requirement of pairwise ranking 
their time complexity is higher than the score-based alternatives. Such a disadvantage of ranking-based loss functions prevents their applicability to larger visual detectors (i.e., transformer-based object detectors). 

\section{Background on Ranking-based Losses}

Ranking-based losses for object detection rely on pairwise comparisons between the scores of different detections to determine the rank of a detection among positives and negatives. Denoting the score of the $i^{th}$ detection by $s_i$, we can compare the scores of two detections $i$ and $j$ with a simple difference transform: $x_{ij} = s_j - s_i$. By counting the number of instances with $x_{ij}>0$, we can determine $i^{th}$ detection's rank among positives ($\mathcal{P}$) and all detections (positives and negatives: $\mathcal{P}\cup\mathcal{N}$) as follows:
\begin{align}\scriptsize
    \operatorname{rank}^{+}(i) = \sum_{j\in \mathcal{P}} \bar{H}(x_{ij}), 
    \text{ and }
    \operatorname{rank}(i) = \sum_{j\in \mathcal{P}\cup\mathcal{N}} \bar{H}(x_{ij})\label{eqn:rank},
\end{align}
where $\bar{H}(x)$ is one if $x>0$ and zero otherwise. Since $\text{d}\bar{H}(x)/\text{d}x$ is either infinite (at $x=0$) or zero (for $x\neq 0$), a smoothed version is used ($\delta$: a hyper-parameter):
\begin{equation}\scriptsize
\label{eq:h_smooth}
\small
H(x)=\left\{
\begin{aligned}
&0\, , & x < -\delta \\
&\frac{x}{2\delta}+0.5\, , & -\delta \leq x \leq \delta \\
&1\, , & \delta < x
\end{aligned}
\right.
\end{equation}

\subsection{Revisiting Average Precision (AP) Loss}
\label{sect:ap_loss}
Given the definitions in Eq. \ref{eqn:rank}, AP Loss \cite{APLoss1,APLoss2} can be defined as:
\begin{equation}\label{eq:ap_loss1}
\small
\mathcal{L}_{AP} = 1-\mathrm{AP}=1-\frac{1}{|\mathcal{P}|} \sum_{i \in \mathcal{P}} \text{precision}(i) =1-\frac{1}{|\mathcal{P}|} \sum_{i \in \mathcal{P}} \frac{\operatorname{rank}^{+}(i)}{\operatorname{rank}(i)}. 
\end{equation}
Chen et al. \cite{APLoss1,APLoss2} simplified Eq. \ref{eq:ap_loss1} by rewriting it in terms of pairwise comparisons between positive and negative detections:
\begin{equation}\label{eq:ap_loss2}
\small
\mathcal{L}_{AP} = \frac{1}{|\mathcal{P}|}\sum_{i\in \mathcal{P}} \sum_{j\in \mathcal{N}} \frac{H(x_{ij})}{\operatorname{rank}(i)} = \frac{1}{|\mathcal{P}|}\sum_{i\in \mathcal{P}} \sum_{j\in \mathcal{N}} L^{AP}_{ij}, 
\end{equation}
{\noindent}where $L^{AP}_{ij}={H(x_{ij})}/{\operatorname{rank}(i)}$ is called the primary term. Note that $ L^{AP}_{ij}$ is non-differentiable since the step function ($H(\cdot)$) applied on $x_{ij}$ is non-differentiable, which will be discussed next. 

\subsection{Revisiting Identity Update}
Oksuz et al. \cite{RSLoss} showed that various ranking-based losses (including AP Loss in Eqs. \ref{eq:ap_loss1} \& \ref{eq:ap_loss2}) can be written in a general form as:
\begin{align}\scriptsize
\label{eq:LossIdentityUpdate}
\mathcal{L}= \frac{1}{Z}\sum \limits_{i \in \mathcal{P} \cup \mathcal{N}} (\ell(i) - \ell^*(i)), 
\end{align}
where $\ell(i)$ is the ranking-based error (e.g., precision error) computed on the $i$th detection, $\ell^*(i)$ is the target ranking-based error (the lowest error possible) and $Z$ is the normalization constant.
%
%
%
The loss in Eq. \ref{eq:LossIdentityUpdate} can be computed and optimized as follows \cite{RSLoss}: 
%
%

\noindent\textbf{1. Computation of the Loss.} First, each pair of logits ($s_i$ and $s_j$) are compared by calculating their difference transforms as $x_{ij} = s_j - s_i$. With the step function $H(\cdot)$ (Eq. \ref{eq:h_smooth}), the number of detections higher than $s_i$ (and therefore its precision error, $\ell(i)$) can be easily calculated (see Sect. \ref{sect:ap_loss}). Given  $x_{ij}$, the loss can be re-written and calculated in terms of  
primary terms $L_{ij}$ as 
$\mathcal{L} =\frac{1}{Z}\sum \limits_{i \in \mathcal{P} \cup \mathcal{N}}\sum \limits_{j \in \mathcal{P} \cup \mathcal{N} }  L_{ij}$, by taking:
\begin{align}\scriptsize
    \label{eq:PrimaryTermIdentityUpdate}
    L_{ij} = \left(\ell(i) - \ell^*(i) \right) p(j|i),
\end{align}
where $p(j|i)$ is a probability mass function (pmf) that distributes the error computed on the $i$th example over the $j$th example in order to determine the pairwise primary term $L_{ij}$. $p(j|i)$ is commonly taken as a uniform distribution \cite{aLRPLoss,RSLoss}. 

%

\noindent\textbf{2. Optimization of the Loss.} The gradient of the primary term wrt. to the difference transform ($\partial L_{ij}/\partial x_{ij}$) is non-differentiable. Denoting this term by $\Delta x_{ij}$, we have \cite{APLoss1}:
\begin{equation}\scriptsize
    \label{eq:Gradients}
    \frac{\partial \mathcal{L}}{\partial s_i} 
    = \sum \limits_{j,k} \frac{\partial \mathcal{L}}{\partial L_{jk}} \Delta x_{jk} \frac{\partial x_{jk}}{\partial s_i} 
    = \frac{1}{Z} \Big( \sum \limits_{j} \Delta x_{ji} - \sum \limits_{j} \Delta x_{ij} \Big). 
\end{equation}
Therefore, optimizing a ranking-based loss function reduces to determining $\Delta x_{ij}$. Inspired by Chen et al. \cite{APLoss1}, Oksuz et al. \cite{RSLoss} employ Perceptron Learning \cite{Rosenblatt} and show that $\Delta x_{ij}$ in Eq. \ref{eq:Gradients} is simply the primary term itself: $\Delta x_{ij}=-(L^*_{ij}-L_{ij})=-(0-L_{ij})=L_{ij}$, hence the name Identity Update. Plugging this into Eq. \ref{eq:Gradients} yields (see Oksuz et al. \cite{RSLoss} for the steps of the derivation):
\begin{equation}\scriptsize
    \label{eq:Final_Gradients}
    \frac{\partial \mathcal{L}}{\partial s_i} 
    = \frac{1}{Z} \Big( \sum \limits_{j} L_{ji} - \sum \limits_{j} L_{ij} \Big). 
\end{equation}
\textit{Therefore, both computing and optimizing the loss reduces determining the primary term $L_{ij}$}.

\subsection{Revisiting AP Loss and RS Loss with Identity Update}
\label{sect:ap_rs_loss}

\noindent\textbf{AP Loss} \cite{APLoss1,APLoss2}. 
%
%
For defining AP Loss with identity update, the errors in Eq. \ref{eq:LossIdentityUpdate} can be derived as $\ell_{\mathrm{R}}(i)= \frac{N_\mathrm{FP}(i)}{\operatorname{rank}(i)}$, $\ell^*_{\mathrm{R}}(i)=0$ and $Z=|\mathcal{P}|$.
Furthermore, defining $p_{R}(j|i)$ as a uniform pmf, that is $p_{R}(j|i) = \frac{{H}(x_{ij})}{{N_\mathrm{FP}}(i)}$,  we can obtain the primary terms of AP Loss using Identity Update, completing the derivation for computation and optimization (Eq. \ref{eq:Final_Gradients}):
\begin{align}\scriptsize
    \label{eq:APPrimaryTermDefinition}
    L^{AP}_{ij} = \begin{cases} \left(\ell_{\mathrm{R}}(i) - \ell^*_{\mathrm{R}}(i) \right) p_{R}(j|i), & \mathrm{for}\; i \in \mathcal{P}, j \in \mathcal{N} \\
    0, & \mathrm{otherwise},
    \end{cases}
\end{align}

\noindent\textbf{RS Loss} \cite{RSLoss}.  
RS Loss includes an additional sorting objective ($\ell_S(i)$) to promote better-localized positives to be ranked higher than other positives: 
\begin{align}\scriptsize
\label{eq:RSLoss}
    \mathcal{L}_\mathrm{RS} =\frac{1}{|\mathcal{P}|}\sum \limits_{i \in \mathcal{P}}  [ \underbrace{(\ell_{\mathrm{R}}(i)-\ell^*_{\mathrm{R}}(i))}_{\text{ranking}}
    +\underbrace{(\ell_{\mathrm{S}}(i)-\ell^*_{\mathrm{S}}(i))}_{\text{sorting}}].
\end{align}
The primary terms of RS Loss can be obtained in a similar manner to AP Loss (see the Supp.Mat. for derivations), yielding: 
\begin{align}\scriptsize\label{eq:RSPrimaryTermDefinition}
        L^{RS}_{ij} = \begin{cases} \left(\ell_{\mathrm{R}}(i) - \ell^*_{\mathrm{R}}(i) \right) p_{R}(j|i), & \mathrm{for}\; i \in \mathcal{P}, j \in \mathcal{N} \\
        \left(\ell_{\mathrm{S}}(i) - \ell^*_{\mathrm{S}}(i) \right) p_{S}(j|i), & \mathrm{for}\;i \in \mathcal{P}, j \in \mathcal{P},\\
        0, & \mathrm{otherwise},
        \end{cases}
\end{align}
where 
$p_{S}(j|i)$ is the sorting pmf. Note that RS Loss has the same primary terms with AP Loss for  $i \in \mathcal{P}, j \in \mathcal{N}$. The primary terms for $i \in \mathcal{P}, j \in \mathcal{P}$ directly targets promoting the better-localized positives.

\subsection{Complexity of Ranking-based Loss Functions} 
These loss functions originally have the space complexity of \(\mathcal{O}((|\mathcal{P}|+|\mathcal{N}|)^2)\) \cite{APLoss1,APLoss2} as they need to compare each pair. This quadratic space complexity makes it infeasible compute these loss functions using modern GPUs as the number of logits in object detection is very large. To alleviate that, Chen et al.\cite{APLoss1} introduced two tricks at the expense of making the computation inefficient: (1) {Loop on Positives}: By implementing a loop over the positive examples, Chen et al. managed to reduce the time complexity to  \(\mathcal{O}(|\mathcal{P}|(|\mathcal{P}|+|\mathcal{N}|)) \approx \mathcal{O}(|\mathcal{P}||\mathcal{N}|)\) since $\mathcal{N}\gg \mathcal{P}$ and the space complexity to \(\mathcal{O}(|\mathcal{P}|+|\mathcal{N}|)\). (2) {Discard Trivial Negatives}: AP Loss considers negative examples with lower rank than all positives as trivial and disregards them. 
\section{Bucketed Ranking-based Loss Functions for Efficient Optimisation of Object Detectors}

\begin{algorithm}[t]
\scriptsize

\caption{AP Loss and RS Loss algorithms. Grey shows the additional operations of RS Loss compared to AP Loss.}\label{alg:RSAlgorithm}
\begin{algorithmic}[1]
  \Require \(\{s_i\}\), predicted logits and \(\{t_i\}\), corresponding labels
  \Ensure \(\{g_i\}\), Gradient of loss wrt. input 

  \State $\forall i, \,\,\, g_i \gets 0$, $\mathcal{P} \gets \{i \mid t_i=1\}, \,\,\, \mathcal{N} \gets \{i \mid t_i=0\}$
  \State $s_{\text{min}} \gets \min_{i\in \mathcal{P}}\{s_{i}\}$, $\mathcal{\widehat{N}} \gets \{i\in \mathcal{N} \mid s_{i}> s_{\text{min}}-\delta\}$

    \noindent \fcolorbox{red}{white}{%
  \begin{minipage}{\linewidth}

  \For{$i \in P$} \tikzmark{start1}
    \State \(\forall j\in \mathcal{P}\cup\mathcal{\widehat{N}}\), \(x_{ij}=s_j-s_i\)   
    \tikzmark{end1}
    \State Ranking error \( \ell_{\mathrm{R}}(i)={{H}(x_{ij})}/{N_\mathrm{{FP}}(i)} \) and $\ell_{\mathrm{R}}^*(i)=0$
    \State $\forall j \in \mathcal{\widehat{N}}$, $L_{ij} = \ell_{\mathrm{R}}(i) \cdot p_{\mathrm{R}}(j|i)$  \Comment{Eq. \ref{eq:APPrimaryTermDefinition}} 
    \State \textcolor{gray}{$\forall j \in \mathcal{P}$, Current sorting error \( \ell_{\mathrm{S}}(j) \)} \Comment{Eq. \ref{eq:RSCurrent}}
    \State \textcolor{gray}{$\forall j \in \mathcal{P}$, Target sorting error \( \ell_{\mathrm{S}}^{*}(j) \)} \Comment{Eq. \ref{eq:RSTarget}}
    \State \textcolor{gray}{$\forall j \in \mathcal{P}$, $L_{ij} = (\ell_{\mathrm{S}}(i) - \ell^*_{\mathrm{S}}(i)) \cdot  p_{\mathrm{S}}(j|i)$} \Comment{Eq. \ref{eq:RSPrimaryTermDefinition}}
    \State Obtain gradient $g_i$ for $i$th positive \Comment{Eq. \ref{eq:Final_Gradients}}
    \State Obtain gradients $g_j$ for $\forall j \in \mathcal{\widehat{N}}$ \Comment{Eq. \ref{eq:Final_Gradients}}
    
  \EndFor
    \end{minipage}%
}
  \State $\forall i, \,\,\, g_i \gets g_i / |\mathcal{P}|$ \Comment{Normalization}
\end{algorithmic}
\Textbox{start1}{end1}{\textcolor{red}{\it Inefficient Iterative Implementation}}

\end{algorithm}

\begin{algorithm}[hbt!]
\scriptsize
\caption{Bucketed RS and AP Losses. Grey shows the additional operations of BRS Loss compared to BAP Loss.}\label{alg:bucketed_RS}
\begin{algorithmic}[1]
\Require All scores \(\{s_i\}\) and corresponding labels \(\{t_i\}\) 
 \Ensure Gradient of input \(\{g_i\}\), ranking loss \( \ell_{\mathrm{R}} \), sorting loss \( \ell_{\mathrm{S}} \)
\State \label{al2:sort-logits} Sort logits to obtain \(\hat{s}_1\), \(\hat{s}_2\), $....$, \(\hat{s}_{|S|}\).
\State \label{al2:bucket-sorted-logits}Bucket consecutive negative logits to obtain $B_1, ..., B_{|\mathcal{P}+1|}$.
\State \label{al2:difference-terms} $\forall i \in \mathcal{P}, \forall j^b \in \Tilde{\mathcal{N}}$, Calculate  $x^b_{ij}$
\State \label{al2:current-ranking-error} $\forall i\in\mathcal{P}$, Calculate bucketed ranking error $\ell^b_\textrm{R}$ \Comment{Eq. \ref{eq:BRanking_Error}}
\State \label{al2:bucketed-primary-terms} $\forall i \in \mathcal{P}, \forall j^b \in \Tilde{\mathcal{N}}$, Calculate $L^{b}_{ij}$\Comment{Eq. \ref{eq:BRanking_Error_Update}}
\State \label{al2:current-sorting-error}  \textcolor{gray}{Calculate current sorting error $\ell^{b}_{\mathrm{S}}$}  \Comment{Eq. \ref{eq:RSCurrent}} 
\State \label{al2:target-sorting-error} \textcolor{gray}{Calculate target sorting error $\ell^{*b}_{\mathrm{S}}$} \Comment{Eq. \ref{eq:RSTarget}}
\State \label{al2:primary-terms} \textcolor{gray}{$\forall i \in \mathcal{P}, \forall j \in \mathcal{P}$, $L_{ij} = (\ell_{\mathrm{S}}^b(i) - \ell^{*b}_{\mathrm{S}}(i)) \cdot  p_{\mathrm{S}}(j|i)$} \Comment{Eq. \ref{eq:RSPrimaryTermDefinition}}
\State \label{al2:final-gradients-for-positive} $\forall i \in \mathcal{P}$, obtain gradients $g_i$ \Comment{Eq. \ref{eq:Final_Gradients}}
\State \label{al2:final-gradients-for-prototype} $\forall j \in \Tilde{\mathcal{N}}$, find gradients for each prototype negative $g_j^b$ \Comment{Eq. \ref{eq:Final_Gradients}}
\State \label{al2:final-gradients-for-negative} $\forall j \in \Tilde{\mathcal{N}}$, normalize $g_j^b$ by bucket size $b_j$ to obtain $g_i$, $\forall i \in \mathcal{N}$
\State \label{al2:normalize} $\forall i, \,\,\, g_i \gets g_i / |\mathcal{P}|$ \Comment{Normalization}
\end{algorithmic}
\end{algorithm}

Training object detectors with existing ranking losses has several advantages as we presented before, however, they suffer from significantly large training time.
%
%
%
%
We now address this by 
our Bucketing Approach, and propose our Bucketed AP and RS Losses.

\subsection{Bucketing Negatives for Efficient Optimisation}
Our main intuition is to bucket the sequential negative examples considering that they will have very similar or equal gradients. Fig. \ref{fig:teaser}(a)  demonstrates this phenomenon in which \(n_1, n_2\) both are assigned equal gradients as well as each negative among \(n_3, n_4, n_5, n_6\) in the standard AP Loss. Formally, we sort all given logits \( s_i \in \mathcal{P} \cup \mathcal{N} \) using a conventional sorting methods. Let us denote this sorted permutation of the logits by \(\hat{s}_1\), \(\hat{s}_2\), $....$, \(\hat{s}_{|\mathcal{P} \cup \mathcal{N}|}\), i.e.,  \(\hat{s}_1\) $>$  \(\hat{s}_2\) $>$  $....$ $>$ \(\hat{s}_{\mathcal{|S|}}\).
%
%
%
%
Given these sorted logits and denoting the $i$th positive logit in the ordering by $\hat{s}^+_i$, the buckets of negatives $B_1, ..., B_{|\mathcal{P}+1|}$ can be obtained by:
\begin{align}\label{eq:bucket_order}
    \small
    B_1 > \hat{s}^+_1 > B_2 > \hat{s}^+_2 > ... > \hat{s}^+_{\mathcal{|P|}} > B_{|\mathcal{P}+1|}.
\end{align}
We will denote the size of the $i$th negative bucket by $b_i$. Having created the buckets, we now determine a single logit value for each bucket, which we call as the prototype logit as it is necessary while assigning the ranking error. We denote the prototype logit for the $i^\mathrm{th}$ bucket by $s^b_i$. Note that if $\delta=0$ in Eq. \ref{eq:h_smooth}, then any logit satisfying $\hat{s}^+_{i-1} > s^b_i  > \hat{s}^+_i$ can be a prototype logit as the ranking stays the same. However, in the case of $\delta>0$, the boundaries between the logits are smoothed. That's why, practically, we find it effective to use the mean logit of a bucket as its prototype logit.

Note that this bucketing approach reduces the number of logits (positive and prototype negative) to a maximum of \( 2 |\mathcal{P}| + 1\). As a result, the pairwise errors can now fit into the memory. Consequently, the loop in the red box of Alg. \ref{alg:RSAlgorithm} is no longer necessary, which gives rise to efficient ranking-based loss functions which we discuss in the following.


\subsection{Bucketed Ranking-based Loss Functions}
Here, we introduce Bucketed versions of AP and RS Losses. Please refer to Alg. \ref{alg:bucketed_RS} for the algorithm.

\subsubsection{Bucketed AP Loss} 
\textbf{Definition.} To define Bucketed version of AP Loss, we need to define current ($\ell^{b}_{\mathrm{R}}(i)$) and target ranking errors ($\ell^{b, *}_{\mathrm{R}}(i)$) for the $i$th positive. Similar to previous works \cite{APLoss1,APLoss2,RSLoss}, we use a target ranking error of $0$, i.e., $\ell^{b,*}_{\mathrm{R}}(i)=0$. Defining the current ranking error, similar to Eq. \ref{eq:ap_loss1}, requires $\mathrm{rank}(i)$ and $N_\mathrm{FP}(i)$. These two quantities can easily be defined using the bucket size $b_i$ and the prototype logit. Different from the conventional AP Loss, we obtain the pairwise relation ${H}(x^b_{ij})$ using the prototype logit  $s^b_i$. Particularly, with the ordering in Eq. \ref{eq:bucket_order}, $\mathrm{rank}(i)=\sum_{j=1}^i {H}(x_{ij}) + {H}(x^b_{ij}) b_j$ and $N_{\textrm{FP}}(i)= \sum_{j=1}^i {H}(x^b_{ij}) b_j$ where $x^b_{ij}=s^b_j - s_i$. Then, the resulting ranking error computed on $i$th positive is:
%
%
%
%
%
%
%
\begin{equation}\footnotesize
\label{eq:BRanking_Error}
    \ell^{b}_{\mathrm{R}}(i)= \frac{N_{\textrm{FP}}(i)}{\mathrm{rank}(i)} = \frac{\sum_{j=1}^i {H}(x^b_{ij}) b_j}{\sum_{j=1}^i {H}(x_{ij}) + {H}(x^b_{ij}) b_j} , \quad i \in \mathcal{P}.
\end{equation}
\textbf{Optimisation.} Here, we need to define the primary terms, as the product between the error and the pmf following Identity Update \cite{RSLoss}. Unlike the previous work, the weights of each prototype negative are not equal as a bucket includes varying number of negatives. Therefore, we use a weighted pmf while distributing the ranking error over negatives. Formally, if $j$ is the $j$th prototype negative, then we define the pmf as $p(j^b|i)=b_{j} / {N}_\mathrm{FP}(i)$. The resulting primary term between the $i$th positive and $j^b$th negative is then:
\begin{equation}\footnotesize
\label{eq:BRanking_Error_Update}
    L^{b}_{ij}= \frac{\sum_{j=1}^i {H}(x^b_{ij}) b_j}{\sum_{j=1}^i {H}(x_{ij}) + {H}(x^b_{ij}) b_j} \times \frac{b_{j}}{{N}_\mathrm{FP}(i)} , \quad i \in \mathcal{P}, j \in \mathcal{\Tilde{N}}
\end{equation}
where $\Tilde{\mathcal{N}}$ is the set of prototype negatives. However, we need the primary terms for the $i$th negative, not for the prototype ones. Given $L^{b}_{ij}$, the gradients of the prototype negatives and actual negatives can easily be obtained following Identity Update. However, we still need to find the gradients of the actual negatives. To do so, we simply normalise the prototype gradient by its bucket size $b_j$ and distribute the gradients to actual negatives, completing our method. In Theorem \ref{theorem1}, we establish that our Bucketed AP Loss provides exactly the same gradients with AP Loss when $\delta=0$.

\subsubsection{Bucketed RS Loss} 
Given that we obtained Bucketed AP Loss, converting RS Loss into a bucketed form is more straightforward. This is because (i) the primary term and its gradients of RS Loss are equal to those of AP Loss if $i \in \mathcal{P}$, $j \in \mathcal{N}$ and we simply use $L^b_{ij}$ in such cases; and (ii) if $i \in \mathcal{P}$, $j \in \mathcal{P}$, then an additional term is included in RS Loss to estimate the pairwise relations between positives (Eq. \ref{eq:RSPrimaryTermDefinition}), which is not affected by bucketing. 

\subsection{Theoretical Discussion} 
\label{sec:theoretical}
The first theorem ensures that the gradients provided by our loss functions are identical with their conventional counterparts under certain circumstances. The second states that our algorithm is theoretically faster than the conventional AP and RS Losses. 
\begin{theorem}\label{theorem1}
Bucketed AP Loss and Bucketed RS Loss provide exactly the same gradients with AP and RS Losses respectively when $\delta=0$.
\end{theorem}
%
\begin{theorem}\label{theorem2}
Bucketed RS and Bucketed AP Losses have $\mathcal{O}(\max ((|\mathcal{P}|+|\mathcal{N}|) \log(|\mathcal{P}|+|\mathcal{N}|), |\mathcal{P}|^2))$ time complexity.
%
\end{theorem}
As a result, we observe the same performance empirically in less amount of time. Proofs are provided in Supp. Mat.

\section{Bucketed Rank-Sort (BRS) DETR}
\label{sec:brs-detr}
While transformer-based detectors \cite{DETR,DeformableDETR,CoDETR,dai2022updetr,glip} have been providing the best performance in several object detection benchmarks, optimizing such detectors using ranking-based losses based on performance measures has not been investigated. Here, we address this gap by incorporating our BRS Loss into Co-DETR \cite{CoDETR} as the current SOTA detector on the common COCO benchmark \cite{COCO}. 

Co-DETR increases the number and variation of the positive examples in the transformer head by leveraging one-to-many assignment strategies in anchor-based detectors (such as ATSS \cite{ATSS} and Faster R-CNN \cite{FasterRCNN}) as auxiliary heads. As the number of proposals is large in such detectors, they can easily provide more and diverse examples, resulting in better performance of the transformer head. Co-DETR is originally supervised with the following loss function:
\begin{equation}\label{eq:codetr_overall}\footnotesize
    \sum_{l=1}^{L} \left(\Tilde{\mathcal{L}}_{l}^{Dec} + \lambda_1 \sum_{k=1}^{K} \Tilde{\mathcal{L}}^{Aux}_{k,l} + \lambda_2 \sum_{k=1}^{K} \mathcal{L}^{Aux}_{k,l}\right)
    ,
\end{equation}
where $\lambda_1$ and $\lambda_2$ weigh each loss component. The first two components, $\Tilde{\mathcal{L}}_{l}^{Dec}$ and $\Tilde{\mathcal{L}}^{Aux}_{k,l}$, are the losses of the $l^{th}$ decoder layer in the following form:
%
%
\begin{align}\label{eq:codetr_main_loss}
   \lambda_{cls} \mathcal{L}_{cls} + \lambda_{bbox} \mathcal{L}_{bbox} + \lambda_{IoU} \mathcal{L}_{IoU} ,
\end{align}
and \(\mathcal{L}_{cls} \), \(\mathcal{L}_{bbox}\) and \(\mathcal{L}_{IoU}\) are Focal Loss,  L1 Loss and GIoU Loss respectively \cite{CoDETR}. 
$\Tilde{\mathcal{L}}_{l}^{Dec}$ and $\Tilde{\mathcal{L}}^{Aux}_{k,l}$ differ in their input queries. That is, while $\Tilde{\mathcal{L}}_{l}^{Dec}$ follows standard DETR-based models \cite{DETR,DeformableDETR} with one-to-one matching of the queries, $\Tilde{\mathcal{L}}^{Aux}_{k,l}$ is computed based on positive anchors of $k^{th}$ auxiliary head with one-to-many assignment.
And $\mathcal{L}^{Aux}_{k,l}$ is the conventional loss of the $k^{th}$ auxiliary head, e.g., the weighted sum of classification, localisation and centerness losses for ATSS \cite{ATSS}.

In order to align the training and evaluation objectives better, we replace $\Tilde{\mathcal{L}}_{l}^{Dec}$ and $\Tilde{\mathcal{L}}^{Aux}_{k,l}$ in Eq. \ref{eq:codetr_main_loss} by:
\begin{align}\label{eq:codetr_main_loss_new}\scriptsize
   \mathcal{L}_{BRS} + \lambda_{bbox} \mathcal{L}_{bbox} + \lambda_{IoU} \mathcal{L}_{IoU},
\end{align}
%
%
and set $\lambda_{bbox}$ and $\lambda_{IoU}$ dynamically during training to $\mathcal{L}_{BRS}/\mathcal{L}_{bbox}$ and $\mathcal{L}_{BRS}/\mathcal{L}_{IoU}$ to simplify the hyperparameter tuning following Oksuz et al. \cite{RSLoss}. Similarly, we replace the loss functions of the auxiliary ATSS \cite{ATSS} and Faster R-CNN \cite{FasterRCNN} heads ($\mathcal{L}^{Aux}_{k,l}$ in Eq. \ref{eq:codetr_overall}) by our BRS Loss. This modification also aligns with the objective of the auxiliary heads by the performance measure, hence results in more accurate auxiliary heads \cite{RSLoss}. Furthermore, as using BRS Loss with Faster R-CNN does not require sampling thanks to its robustness to imbalance, no limitation is imposed on the number of positives in Faster R-CNN. Hence, the main aim of Co-DETR, that is to introduce more positive examples to the transformer head, is corroborated. As a result, our BRS DETR enables significantly efficient training compared to RS Loss also by improving the performance of Co-DETR.

\section{Experiments}

We analyze our contributions in three primary sections. First, we evaluate the effectiveness of our bucketing approach by contrasting it with  RS Loss and AP Loss across different CNN-based visual detectors, on object detection and instance segmentation tasks. 
Next, we thoroughly examine how our bucketed RS Loss performs on transformer-based object detectors with different backbones. 
This is the first time a ranking-based loss is applied to transformer-based object detectors,  thanks to the time-efficiency of our bucketing method. 
Finally, we present comparisons with score-based loss functions and other ranking-based loss functions in the literature. 
%
Our experiments clearly show that in terms  of training time, accuracy and ease of tuning, our BRS Loss is more preferable over any existing loss function to train object detectors.

\noindent\textbf{Dataset and Performance Measures.} Unless otherwise noted, we train all models on COCO \textit{trainval35k} (115k images) and test them on \textit{minival} (5k images). We use COCO-style Average Precision (AP) and also report $\mathrm{AP_{50}}$, $\mathrm{AP_{75}}$ as the APs at IoUs $0.50$ and $0.75$; and  $\mathrm{AP_{S}}$, $\mathrm{AP_{M}}$ and $\mathrm{AP_{L}}$ to present the accuracy on small, medium and large objects.

\subsection{Efficiency and Effectiveness of Bucketed Ranking Losses}
\label{subsec:brs_vs_rs}

To comprehensively analyze the effectiveness of our bucketing method, we design experiments on real-world data as well as synthetic data. For the former, we aim to present that our method significantly decreases the training time of the detection and segmentation methods (up to $\sim 5 \times$). As training the entire detector does not isolate the runtime of the loss function, on which our main contribution is, we also design an experiment with synthetic data. This set of experiments show that our bucketing approach, in fact, decreases the loss function runtime by up to $40 \times$, thereby resulting in shorter training time of the detectors.

\noindent \textbf{Experimental Setup.}
We follow the experimental setup used in  RSLoss \cite{RSLoss} only by replacing the loss function by our bucketed loss function to ensure comparability and consistency of results.
We report our findings on efficiency and accuracy using the models trained on 4 Tesla A100 GPUs with a batch size of 16, i.e., 4 images/GPU.
Unless stated otherwise, we train the models for 12 epochs and test them on images with a size of 1333$\times$800 following the common convention \cite{ATSS,RSLoss,paa}.
In order to comprehensively demonstrate the efficiency of our approach, we use five different detectors: Faster R-CNN \cite{FasterRCNN}, Cascade R-CNN \cite{CascadeRCNN}, ATSS \cite{ATSS} and PAA \cite{paa} for object detection as well as Mask R-CNN \cite{MaskRCNN} for instance segmentation. 


\begin{table}[t]
\centering
\caption{Bucketed Losses (BRS, BAP) vs. RS \& AP Losses on COCO object detection  for object detectors. R-$X$: ResNet-$X$. Models with R-101 are trained 36 epochs.
\label{tab:MSOD_RSvsBRS}}
\setlength{\tabcolsep}{0.8em}
\scalebox{0.83}{
\begin{tabular}{|c|c|c|c|c|c|c|c|c|}
\hline
\multirow{2}{*}{Method}&\multirow{2}{*}{Detector}&\multirow{2}{*}{BB}&\multirow{2}{*}{$\mathcal{L}$}&\multicolumn{2}{|c|}{Time}&\multicolumn{3}{|c|}{Accuracy}\\ \cline{5-9} 
&&&&\multicolumn{2}{|c|}{$T_{iter}(s) \downarrow$}&$\mathrm{AP} \uparrow$&$\mathrm{AP_{50}} \uparrow$&$\mathrm{AP_{75}} \uparrow$\\ \hline\hline
\multirow{6}{*}{Multi Stage}&\multirow{4}{*}{Faster R-CNN}&\multirow{2}{*}{R-50}&RS&\multicolumn{2}{|c|}{$0.50$\phantom{\scriptsize{ 3.0x $\downarrow$}}}&$39.4$&$59.5$&$43.0$\\ \cdashline{4-9}[.4pt/1pt]
&&&BRS&\multicolumn{2}{|c|}{$\textbf{0.17}$\textcolor{darkspringgreen}{\scriptsize{ 3.0x $\downarrow$}}} &$39.5$&$59.5$&$42.8$\\ \cline{3-9}
&&\multirow{2}{*}{R-101}&RS&\multicolumn{2}{|c|}{$0.75$\phantom{\scriptsize{ 3.0x $\downarrow$}}}&$47.3$&$67.4$&$51.2$\\ \cdashline{4-9}[.4pt/1pt]
&&&BRS&\multicolumn{2}{|c|}{$\textbf{0.38}$\textcolor{darkspringgreen}{\scriptsize{ 2.0x $\downarrow$}}}&$47.7$&$67.8$&$51.5$\\ \cline{2-9}
&\multirow{2}{*}{Cascade R-CNN}&\multirow{2}{*}{R-50}&RS&\multicolumn{2}{|c|}{$1.28$\phantom{\scriptsize{ 5.3x $\downarrow$}}}&$41.1$&$58.7$&$44.1$\\ \cdashline{4-9}[.4pt/1pt]
&&&BRS&\multicolumn{2}{|c|}{$\textbf{0.24}$\textcolor{darkspringgreen}{\scriptsize{ 5.3x $\downarrow$}}}&$41.1$&$58.8$&$44.2$\\ \hline \hline
\multirow{8}{*}{One Stage}&\multirow{4}{*}{ATSS}&\multirow{8}{*}{R-50}
&AP&\multicolumn{2}{|c|}{$0.32$\phantom{\scriptsize{ 2.1x $\downarrow$}}}&$38.1$&$58.2$&$41.0$ \\ \cdashline{4-9}[.4pt/1pt]
&&&BAP&\multicolumn{2}{|c|}{$\textbf{0.15}$\textcolor{darkspringgreen}{\scriptsize{ 2.1x $\downarrow$}}}&$38.5$&$58.6$&$41.3$ \\ \cline{4-9}
&&&RS&\multicolumn{2}{|c|}{$0.36$\phantom{\scriptsize{ 2.3x $\downarrow$}}}&$39.8$&$58.9$&$42.6$\\ \cdashline{4-9}[.4pt/1pt] 
&&&BRS&\multicolumn{2}{|c|}{$\textbf{0.15}$\textcolor{darkspringgreen}{\scriptsize{ 2.4x $\downarrow$}}}&$39.8$&$58.8$&$42.9$\\ \cline{2-2}\cline{4-9}
&\multirow{4}{*}{PAA}&
&AP&\multicolumn{2}{|c|}{$0.45$\phantom{\scriptsize{ 1.5x $\downarrow$}}}&${37.3}$&$54.3$&$41.2$ \\ \cdashline{4-9}[.4pt/1pt]
&&&BAP&\multicolumn{2}{|c|}{$\textbf{0.30}$\textcolor{darkspringgreen}{\scriptsize{ 1.5x $\downarrow$}}}&${37.2}$&$56.2$&$40.2$ \\ \cline{4-9}
&&&RS&\multicolumn{2}{|c|}{$0.57$\phantom{\scriptsize{ 1.9x $\downarrow$}}}&$40.8$&$58.8$&$44.6$\\ \cdashline{4-9}[.4pt/1pt]
&&&BRS&\multicolumn{2}{|c|}{$\textbf{0.30}$\textcolor{darkspringgreen}{\scriptsize{ 1.9x $\downarrow$}}}&$40.9$&$59.0$&$44.4$\\ \hline
\end{tabular}
}
\end{table}

\blockcomment{
\begin{table}[ht]
\centering\setlength{\tabcolsep}{3pt}\scriptsize

\begin{tabular}{|c|c|c|c|c|c|c|c|}
\hline
\multirow{2}{*}{Detector}&\multirow{2}{*}{Backbone}&\multirow{2}{*}{$\mathcal{L}$}&\multicolumn{2}{|c|}{Efficiency}&\multicolumn{3}{|c|}{Accuracy}\\
\cline{4-8} 
&&&\multicolumn{2}{|c|}{$T_{iter}(s) \downarrow$}&$\mathrm{AP} \uparrow$&$\mathrm{AP_{50}} \uparrow$&$\mathrm{AP_{75}} \uparrow$\\ \hline\hline
\multirow{2}{*}{ATSS}&\multirow{2}{*}{R-50}&RS&\multicolumn{2}{|c|}{$0.36$\phantom{\scriptsize{ 2.3x $\downarrow$}}}&$39.8$&$58.9$&$42.6$\\ \cdashline{3-8}[.4pt/1pt] 
&&BRS&\multicolumn{2}{|c|}{$\textbf{0.15}$\textcolor{darkspringgreen}{\scriptsize{ 2.4x $\downarrow$}}}&$39.8$&$58.8$&$42.9$\\ \hline\hline
\multirow{2}{*}{PAA}&\multirow{2}{*}{R-50}&RS&\multicolumn{2}{|c|}{$0.57$\phantom{\scriptsize{ 1.9x $\downarrow$}}}&$40.8$&$58.8$&$44.6$\\ \cdashline{3-8}[.4pt/1pt]
&&BRS&\multicolumn{2}{|c|}{$\textbf{0.30}$\textcolor{darkspringgreen}{\scriptsize{ 1.9x $\downarrow$}}}&$40.9$&$59.0$&$44.4$\\ \hline
\end{tabular}
\caption{Bucketed RS (BRS) Loss vs. RS Loss on COCO object detection for one stage object detectors. R-50: ResNet-50.}
\label{tab:OSOD_RSvsBRS}
\end{table}
}
\noindent\textbf{Bucketed Ranking Based Losses Significantly Decrease the Training Time with Similar Accuracy.}
\textit{Multi-stage detectors}: Among multi-stage detectors, we train the commonly-used Faster R-CNN and Cascade R-CNN with RS Loss and BRS Loss, and present average iteration time as well as AP of the trained models. 
In order to evaluate our contribution comprehensively, we train Faster R-CNN also with a stronger setting, in which, we use ResNet-101, train it for 36 epochs using multi-scale training similar to \cite{RSLoss}.
The results are presented in Table \ref{tab:MSOD_RSvsBRS}, in which we can clearly see that BRS Loss consistently reduces the training time of all three detectors as well as preserves (or slightly improve) their performance.
Especially for Cascade R-CNN, which is still a very popular and strong object detector along with its variants such as HTC \cite{HTC}, the training time decreases by $5.3\times$.
This is because the loss function is applied to the logits for three times based on the cascaded nature of this detector, and therefore, our contribution can be easily noticed.
Moreover, it might seem that the efficiency gain decreases once the size of the backbone increases from R-50 to R-101 in Faster R-CNN.
However, this is an expected result as the overall feature extraction time increases due to the larger number of parameters in the backbone, and hence, as we will discuss in the synthetic experiments, this is independent from our bucketing approach, operating on the loss function after the feature extraction.

\textit{One-stage detectors}: To  show that our gains generalize to one-stage detectors and to AP Loss, 
we train the common ATSS and PAA  detectors with our BAP and BRS Losses, and compare them with AP and RS Losses respectively.
Table \ref{tab:MSOD_RSvsBRS} validates our previous claims: Our bucketed losses obtain similar performance in around half of the training time required for AP and RS Losses.
%
%

\textit{Instance Segmentation:} Given that our loss functions is efficient in object detectors, one simple extension is to see their generalization to instance segmentation methods.
To show that, we train Mask R-CNN, a common baseline, with our BRS Loss on three different dataset from various domains: (i) COCO (Table \ref{tab:IS_RSvsBRS}), (ii) Cityscapes \cite{Cordts2016Cityscapes} as an autonomous driving dataset (Table \ref{tab:additional_datasets}) and LVIS \cite{LVIS} a long-tailed dataset with more than 1K classes (Table \ref{tab:additional_datasets}).
The results confirm our previous finding that  our BRS Loss decreases the training time significantly by up to $\sim 2.5 \times$ compared to RS Loss by preserving its accuracy.

\begin{table}[t]
\parbox{.48\linewidth}{
\setlength{\tabcolsep}{0.4em}
\centering
\caption{Comparison with RS Loss on instance segmentation task on COCO val using Mask R-CNN. BB: Backbone.
}
\scalebox{0.77}{
\begin{tabular}{|c|c|c|c|c|c|c|}
\hline
\multirow{2}{*}{BB}&\multirow{2}{*}{$\mathcal{L}$}&\multicolumn{2}{|c|}{Efficiency}&\multicolumn{3}{|c|}{Accuracy}\\ \cline{3-7} 
&& \multicolumn{2}{|c|}{$T_{iter}(s) \downarrow$}&$\mathrm{AP} \uparrow$&$\mathrm{AP_{50}} \uparrow$ & $\mathrm{AP_{75}} \uparrow$\\ \hline \hline
\multirow{2}{*}{R-50}&RS&\multicolumn{2}{|c|}{$0.56$\scriptsize{\phantom{2.3x $\downarrow$}}}&$36.3$&$57.2$&$38.8$\\ \cdashline{2-7}[.4pt/1pt]
&BRS&\multicolumn{2}{|c|}{$\textbf{0.24}$\textcolor{darkspringgreen}{\scriptsize{ 2.3x $\downarrow$}}}&$36.2$&$57.2$&$38.8$\\ \hline
\multirow{2}{*}{R-101}&RS&\multicolumn{2}{|c|}{$0.59$\phantom{\scriptsize{ 2.1x $\downarrow$}}}&$40.2$&$61.8$&$43.5$\\ \cdashline{2-7}[.4pt/1pt]
&BRS&\multicolumn{2}{|c|}{$\textbf{0.27}$\textcolor{darkspringgreen}{\scriptsize{ 2.2x $\downarrow$}}}&$40.3$&$62.0$&$43.8$\\ \hline
\end{tabular}
\label{tab:IS_RSvsBRS}
}
}
\hfill
\parbox{.48\linewidth}{
\small
\setlength{\tabcolsep}{0.3em}
\centering
\caption{Comparison on different inst. segm. datasets using Mask R-CNN. $\mathrm{AP_{75}}$ is N/A as it is not used for Cityscapes.
}
\scalebox{0.77}{
\begin{tabular}{|c|c|c|c|c|c|c|}
\hline
\multirow{2}{*}{Dataset}&\multirow{2}{*}{$\mathcal{L}$}&\multicolumn{2}{|c|}{Efficiency}&\multicolumn{3}{|c|}{Accuracy}\\ \cline{3-7} 
&& \multicolumn{2}{|c|}{$T_{iter}(s) \downarrow$}&$\mathrm{AP} \uparrow$&$\mathrm{AP_{50}} \uparrow$ & $\mathrm{AP_{75}} \uparrow$\\ \hline \hline
\multirow{2}{*}{Cityscapes}&RS&\multicolumn{2}{|c|}{$0.43$\phantom{\scriptsize{ 2.3x $\downarrow$}}}&$43.5$&$68.1$ & N/A\\ \cdashline{2-7}[.4pt/1pt]
&BRS&\multicolumn{2}{|c|}{\textbf{0.19}\textcolor{darkspringgreen}{\scriptsize{ 2.3x $\downarrow$}}}&$43.3$&$67.7$&N/A\\ \hline
\multirow{2}{*}{LVIS}&RS&\multicolumn{2}{|c|}{$0.87$\phantom{\scriptsize{ 2.5x $\downarrow$}}}&$25.6$&$39.2$ & $27.3$ \\ \cdashline{2-7}[.4pt/1pt] 
&BRS&\multicolumn{2}{|c|}{\textbf{0.35}\textcolor{darkspringgreen}{\scriptsize{ 2.5x $\downarrow$}}}&$25.8$&$39.6$&$27.4$\\ \hline 
\end{tabular}
\label{tab:additional_datasets}
}
}
\end{table}

\begin{table}[t]
\caption{BRS Loss further improves performance with similar training budget. We report mask AP for instance segmentation}
\label{tab:BRS_with_same_time_cascade}
\centering\setlength{\tabcolsep}{2pt}\scriptsize
\begin{tabular}{|c|c|c|c|c|c|}
\hline
Task&Detector&$\mathcal{L}$&$\mathrm{AP} \uparrow$&$\mathrm{AP_{50}} \uparrow$&$\mathrm{AP_{75}} \uparrow$ \\ \hline \hline

\multirow{2}{*}{Object Det.}&\multirow{2}{*}{C. R-CNN}&RS&$41.1$\phantom{\scriptsize{ +1.2}}&$58.7$&$44.1$\\\cdashline{3-6}[.4pt/1pt]
&&BRS&\textbf{42.3}\textcolor{darkspringgreen}{\scriptsize{ +1.2}}&\textbf{60.2}&\textbf{45.2}\\\hline \hline
\multirow{2}{*}{Inst. Seg.}&\multirow{2}{*}{M. R-CNN}&RS&$36.3$\phantom{\scriptsize{ +1.0}}&$57.2$&$38.8$\\\cdashline{3-6}[.4pt/1pt]
&&BRS&\textbf{37.3}\textcolor{darkspringgreen}{\scriptsize{ +1.0}}&\textbf{58.4}&\textbf{40.2}\\\hline
\end{tabular}
\end{table}

\noindent\textbf{Given Similar Training Budget, Bucketed Ranking Based Loss Functions Significantly Improve the Accuracy.} There might be several benefits of reducing the training time of the detector as more GPU time is saved. 
Here, we show a use-case in which we ask the following question: what would happen if we allocate the similar amount of training time to both bucketed and non-bucketed losses? To answer that, we train Cascade R-CNN, a detector, and Mask R-CNN, an instance segmentation method, using our BRS Loss. 
Considering our speed-ups on these models ($6.0 \times$ and $2.3 \times$), we simply increased their training epochs from 12 to 36 and 27 respectively.
Note that, for Cascade R-CNN, we still spend  significantly less amount of training time.
Table \ref{tab:BRS_with_same_time_cascade} shows that models trained with the BRS Loss outperform (with $+1.2$ and $1.0$ AP)  their counterparts trained with the RS Loss in both cases.




\noindent\textbf{Experiments on Synthetic Data.}
%
Up to now, to measure any speed-up, we considered the entire pipeline and did not decouple the loss computation step. 
%
However, we only improve the  loss computation step. Hence, an analysis focusing only on this part can reveal our contribution more clearly.
Therefore,  we design a  simple  experiment using synthetic data.
Particularly, we randomly generate $L$ logits such that $m \%$ of these logits are positive (see Supp.Mat. for data generation details).
To cover the wide range of existing detectors, we set $L \in \{10K, 100K, 1M\}$ and $m \in \{0.1, 1.0, 2.0, 5.0\}$.
Different choices for $L$ and $m$ correspond to the logit distributions of common object detectors.
%
%
%
We compute BAP and AP Loss on this synthetic data and compare their runtime to focus on loss computation time.
We observed that our BAP Loss has a significant, up to $\sim 40\times$, speed-up compared to AP Loss. 
The speed-up becomes more pronounced with increasing number of logits. Supp. Mat. presents detailed results.

\begin{table}[t]
\parbox{.48\linewidth}{
\setlength{\tabcolsep}{0.2em}
\centering
\caption{Comparison of our BRS-DETR
with Co-DETR (trained with its original loss function) on different backbones on COCO val set.
}
\scalebox{0.7}{
\begin{tabular}{|c|c||c|c|c|c|c|c|}
\hline
Backbone&Detector&$\mathrm{AP}$&$\mathrm{AP_{50}}$&$\mathrm{AP_{75}}$&$\mathrm{AP_{s}}$&$\mathrm{AP_{m}}$&$\mathrm{AP_{l}}$ \\ \hline \hline
\multirow{2}{*}{ResNet50}&Co-DETR & $49.3$ & $67.2$ & $54.0$ & $\textbf{32.1}$ & $52.6$ & $63.8$ \\ \cdashline{2-8}[.4pt/1pt]
& BRS-DETR& $\textbf{50.1}$& $\textbf{67.4}$ & $\textbf{54.6}$ & $31.9$ & $\textbf{53.9}$ & $\textbf{65.0}$ \\ \hline \hline
\multirow{2}{*}{Swin-T}&Co-DETR&$51.7$&$\textbf{69.6}$&$56.4$&$34.4$&$54.9$&$66.8$ \\ \cdashline{2-8}[.4pt/1pt]
& BRS-DETR&$\textbf{52.3}$ &$69.5$&$\textbf{57.1}$&$\textbf{34.6}$&$\textbf{55.7}$&$\textbf{68.0}$ \\ \hline \hline
\multirow{2}{*}{Swin-L}&Co-DETR&$56.9$&$\textbf{75.5}$&$\textbf{62.6}$&$\textbf{40.1}$&$61.2$&$73.3$ \\ \cdashline{2-8}[.4pt/1pt]
& BRS-DETR&\textbf{57.2}& 75.0& 62.5& 39.4& \textbf{61.7}& \textbf{74.1}\\ \hline
\end{tabular}
\label{tab:detr_backbone_comparison}
}
}
\hfill
\parbox{.48\linewidth}{
\small
\setlength{\tabcolsep}{0.2em}
\centering
\caption{Comparison of our BRS-DETR with DETR variants (trained with their original loss functions) w ResNet-50 on COCO val set. 
Qu: Queries. Ep: Epochs.
}
\scalebox{0.68}{
\begin{tabular}{|c|c|c||c|c|c|c|c|c|}
\hline
Detector&Qu&Ep&$\mathrm{AP}$&$\mathrm{AP_{50}}$&$\mathrm{AP_{75}}$&$\mathrm{AP_{s}}$&$\mathrm{AP_{m}}$&$\mathrm{AP_{l}}$ \\ \hline \hline 
DETR \cite{DETR} & $100$ & $300$ & $42.0$ & $62.4$ & $44.2$ & $20.5$ & $45.8$ & $61.1$ \\ \hline
DN-DETR \cite{dn-detr}  & $300$ & $50$ & $48.6$ & $67.4$ & $52.7$ & $31.0$ & $52.0$ & $63.7$ \\ \hline
DINO \cite{DINO} & $900$ & $12$ & $49.4$ & $66.9$ & $53.8$ & $32.3$ & $52.5$ & $63.9$ \\ \hline
H-DETR \cite{H-DETR} & $300$ & $12$ & $48.7$ & $66.4$ & $52.9$ & $31.2$ & $51.5$ & $63.5$ \\ \hline
Co-DETR\cite{CoDETR} & $300$ & $12$ & $49.3$ & $67.2$ & $54.0$ & $32.1$ & $52.6$ & $63.8$ \\ \hline \hline
BRS-DETR & $300$ & $12$ & $\textbf{50.1}$ & $67.4$ & $\textbf{54.6}$ & $31.9$ & $\textbf{53.9}$ & $\textbf{65.0}$ \\ \hline

\end{tabular}
\label{tab:transformer_comparison}
}
}
\end{table}

\subsection{The Effectiveness of Our BRS-DETR}
While we incorporate our BRS Loss into Co-DETR, we 
 use all original settings unless otherwise noted. Please refer to Supp.Mat. for the details.
 

\noindent\textbf{BRS-DETR Outperforms Existing Transformer-based Detectors Consistently.} 
Here, we first employ ResNet-50 backbone as it has been used by many detectors, enabling us to compare our method with many different DETR-based manner in a consistent manner. In this fair comparison setting, our BRS-DETR outperforms all existing DETR variants reaching $50.1$ AP as shown in Table \ref{tab:transformer_comparison}. For example, our BRS DETR outperforms DN-DETR by $1.5$AP with less training epochs and improves Co-DETR as we discuss next.

\noindent\textbf{Training Co-DETR with RS Loss takes 6x less time compared to using RS Loss. 
} Finally, we also compare the training efficiencies of BRS Loss and RS Loss on Co-DETR. We observe that our BRS Loss decreases the training time of Co-DETR by $6.0 \times$ (from $4.14$s per iteration to $0.69$s) in comparison to RS Loss. We note that this is the largest training time gain of our BRS Loss. This is because Co-DETR consists of multiple transformer-based as well as auxiliary heads, requiring multiple loss estimations.

\noindent\textbf{BRS-DETR Improves Co-DETR over Different Backbones Consistently.}
In order to show the effectiveness of our BRS Loss, we compare our results with Co-DETR in Table \ref{tab:detr_backbone_comparison} using different backbones. We note that we improve the baseline Co-DETR consistently in all settings. For example, our improvement on ResNet-50 is $0.8$AP, which is a notable improvement. However, as the backbone gets larger (e.g., Swin-L \cite{SwinTransformer}), our gains decrease, which is expected and commonly observed in the literature, e.g., \cite{paa}.

\subsection{Comparison with Score-based and Other Ranking-based Losses 
}
In previous sections, we showed the efficiency of our loss functions compared to AP and RS Losses as their counterparts.
Here, we compare our bucketed losses with the score based losses, i.e., Cross-entropy and Focal Loss, and other ranking based losses including DR Loss \cite{DRLoss} and aLRP Loss \cite{aLRPLoss}. 
%
To do so, in Tables \ref{tab:RBvsSB} and \ref{tab:RBvsotheranking}, we report average training time, AP for accuracy and number of hyperparameters to capture the tuning simplicity of the loss functions.
Compared to score-based loss functions in Table \ref{tab:RBvsSB}, our BRS Loss improves performance of both approaches, is significantly simpler-to-tune with a low number of hyperparameters and has similar training iteration time.
Hence, our BRS Loss is either superior or on par with existing score-based losses.
As for other ranking-based losses in Table \ref{tab:RBvsotheranking}, our BRS Loss is the most efficient and accurate ranking-based loss function for all three detectors.
As an example, compared to ATSS trained with DR Loss, our BRS Loss yields $1.8$AP better accuracy with $25\%$ less training time and significantly less number of hyperparameters. 
Therefore, our bucketed loss functions are now promising alternatives to train object detectors.


\begin{table}[t]

\parbox{.48\linewidth}{
\setlength{\tabcolsep}{0.3em}
\centering
\caption{Comparison with the score-based losses. 
CE: Cross-entropy. \#H: Number of hyperparameters. Ours has very similar $T_{iter}(s)$ also by being more accurate and simple-to-tune.
}
\scalebox{0.77}{
\begin{tabular}{|c|c|c|c|c|}

\hline
Detector&$\mathcal{L}$&$T_{iter}(s) \downarrow$&$\mathrm{AP} \uparrow$ &\#H \\ \hline \hline
\multirow{3}{*}{Faster R-CNN}&CE+L1&\textbf{0.14}&$37.6$&$9$ \\\cdashline{2-5}[.4pt/1pt]
&RS&$0.50$&$39.4$&\textbf{3} \\\cdashline{2-5}[.4pt/1pt]
&BRS (Ours)&$0.17$&\textbf{39.5}&\textbf{3} \\\hline \hline
\multirow{3}{*}{ATSS}&Focal L.+GIoU&\textbf{0.14}&$39.3$&$5$ \\\cdashline{2-5}[.4pt/1pt]
&RS&$0.36$&\textbf{39.8}&\textbf{1} \\\cdashline{2-5}[.4pt/1pt]
&BRS (Ours)&$0.15$&\textbf{39.8}&\textbf{1} \\\hline

\end{tabular}
\label{tab:RBvsSB}
}
}
\hfill
\parbox{.48\linewidth}{
\small
\setlength{\tabcolsep}{0.35em}
\centering
\caption{Comparison with other ranking-based losses. Our approach performs better on both accuracy and efficiency.
}
\scalebox{0.77}{
\begin{tabular}{|c|c|c|c|c|}
\hline
Detector&$\mathcal{L}$&$T_{iter}(s) \downarrow$&$\mathrm{AP} \uparrow$ &\#H \\ \hline \hline
\multirow{2}{*}{Faster R-CNN}&aLRP \cite{aLRPLoss}&$0.28$&$37.4$&\textbf{3} \\\cdashline{2-5}[.4pt/1pt]
&BRS (Ours)&\textbf{0.17}&\textbf{39.5}&\textbf{3} \\\hline \hline
\multirow{4}{*}{ATSS}&AP \cite{APLoss1}&$0.32$&$38.1$&$5$ \\\cdashline{2-5}[.4pt/1pt]
&DR \cite{DRLoss} &$0.20$&$38.1$& 5 \\\cdashline{2-5} [.4pt/1pt]
&aLRP \cite{aLRPLoss}&$0.32$&$37.7$&\textbf{1} \\\cdashline{2-5} [.4pt/1pt]
&BRS (Ours)&\textbf{0.15}&\textbf{39.8}&\textbf{1} \\\hline \hline
\multirow{2}{*}{RetinaNet} & DR \cite{DRLoss} &$0.29$&$37.4$&5\\  \cdashline{2-5}[.4pt/1pt]
& BRS (Ours)  & $\textbf{0.23}$&$\textbf{38.3}$& \textbf{1}\\ \hline
\end{tabular}
\label{tab:RBvsotheranking}
}
}
\end{table}

\section{Conclusion}

In this paper, we introduced a novel method to improve the efficiency of ranking-based loss functions by binning negatives into buckets and implementing positive-negative pairwise comparisons between positives and buckets of negatives. We showed that this method improves the computational complexity to a level where pairwise comparisons can be stored as a matrix in memory and running time becomes very close to that of score-based loss functions. Also, for the first time,  we integrated a ranking-based loss to Co-DETR, a DETR-based detector, which was possible thanks to our method's lower complexity, and reported improvements on different backbones.  
Our comprehensive experiments comprising 2 different tasks, 3 different dataset and 7 different detectors -- all with positive outcomes -- show the general applicability of our approach.

\section*{Acknowledgements}
We gratefully acknowledge the computational resources provided by T\"UB\.ITAK ULAKB\.IM High Performance and Grid Computing Center (TRUBA), METU Center for Robotics and Artificial Intelligence (METU-ROMER) and METU Image Processing Laboratory. Dr. Akbas is supported by the ``Young Scientist Awards Program (BAGEP)'' of Science Academy, Türkiye.

{
    \small
    \bibliographystyle{splncs04}
    \bibliography{main}
}

\newpage
\setcounter{section}{0}
\setcounter{equation}{0}
\setcounter{figure}{0}
\setcounter{table}{0}
\renewcommand{\thefigure}{S.\arabic{figure}}
\renewcommand{\thetable}{S.\arabic{table}}
\renewcommand{\theequation}{S.\arabic{equation}}
\renewcommand\thesection{S.\arabic{section}}
\renewcommand\thesubsection{\thesection.\arabic{subsection}}
\section*{Supplementary Material for ``Bucketed Ranking-based Losses for Efficient Training of Object Detectors''}

\startcontents
\printcontents{}{1}{\section*{Contents}}

\section{Further Details of Existing Ranking Based Losses}
For the sake of completeness, this section presents further details of the existing ranking-based loss functions, AP \cite{APLoss1,APLoss2} and RS Losses \cite{RSLoss}, which are excluded in the paper due to the space limitation.
\subsection{Further Details of AP Loss}
We presented in Section 3.3 almost a complete derivation of AP Loss except including its gradients for the $i$th logit, that is, $\frac{\partial\mathcal{L}_{AP}}{\partial s_i}$. 
The main reason why we left the gradients out is that $\frac{\partial\mathcal{L}_{AP}}{\partial s_i}$ (or also for RS Loss) is trivial to obtain once primary terms are derived.
To do so, we simply replace the primary terms ($L_{ij}$ in Eq. 8)  into the gradient definition of Identity Update and set $Z=|\mathcal{P}|$ in Eq. 5.
Using mathematical simplifications, this yields the following gradients for the $i$th logit for AP Loss:
\begin{align}
    \label{eq:APGradient}
    \frac{\partial\mathcal{L}_{AP}}{\partial s_i} = \begin{cases}      \frac{1}{|\mathcal{P}|} \sum_{j \in \mathcal{P}} \ell_R(j) p_R(i|j), & \mathrm{for}\; i \in \mathcal{N} \\
    -\frac{1}{|\mathcal{P}|} \ell_{\mathrm{R}}(i), & \mathrm{for}\; i \in \mathcal{P},
    \end{cases}
\end{align}

\subsection{Further Details of RS Loss}
This section presents further details of RS Loss, in which we define the loss function as well as derive its gradients based on Identity Update.
RS Loss is the average difference between the current ($\ell_{\mathrm{RS}}(i)$) and the target ($\ell_{\mathrm{RS}}^*(i)$) RS errors over positives where labels for positives are defined as their IoU values, $IoU_i$:
\begin{align}
\label{eq:s_RSLoss}
    \mathcal{L}_\mathrm{RS}:=\frac{1}{|\mathcal{P}|}\sum \limits_{i \in \mathcal{P}}  \left( \ell_{\mathrm{RS}}(i)-\ell_{\mathrm{RS}}^*(i) \right),
\end{align}
where $\ell_{\mathrm{RS}}(i)$ is a summation of the current ranking error and current sorting error and ($\ell_{\mathrm{RS}}^*(i)$) is a summation  of the target ranking error and target sorting error:
\begin{align}
\label{eq:RSCurrent}
    \ell_{\mathrm{RS}}(i):= \underbrace{ \frac{\mathrm{N_{FP}}(i)}{\mathrm{rank}(i)}}_{\ell_{\mathrm{R}}(i):\ \text{Current Ranking Error}}
    +\underbrace{\frac{\sum \limits_{j \in \mathcal{P}} \mathrm{H}(x_{ij})(1-IoU_j)}{\mathrm{rank^+}(i)}}_{\ell_{\mathrm{S}}(i):\ \text{Current Sorting Error}}. 
\end{align}
When $i$ belongs to the positive set $\mathcal{P}$, the current ranking error is calculated as the precision error. The current sorting error, on the other hand, assigns a penalty to the positives with logits greater than $s_i$. This penalty is proportional to the average of their inverted labels, $1-IoU_j$:
\begin{align}
\label{eq:RSTarget}
    \ell^*_{\mathrm{RS}}(i)= \cancelto{0}{\ell^*_{\mathrm{R}}(i)}+ \underbrace{\frac{\sum \limits_{j \in \mathcal{P}} \mathrm{H}(x_{ij})[IoU_j \geq IoU_i](1-IoU_j)}{\sum \limits_{j \in \mathcal{P}} \mathrm{H}(x_{ij})[IoU_j \geq IoU_i]}}_{\ell^*_{\mathrm{S}}(i):\text{Target Sorting Error}},
\end{align}
where $[.]$ denotes the Iverson Bracket. The target ranking of $i$, which is based on its desired position in the ranking, is compared to this measure. The target sorting error is calculated by averaging over the inverted IoU values of $j \in \mathcal{P}$ with larger logits and labels than $i \in \mathcal{P}$, corresponding to the desired sorted order.

%

Based on these definitions, we presented the following primary terms of RS Loss in Section 3.3:
\begin{align}\footnotesize
    \label{eq:s_RSPrimaryTermDefinition}
    L_{ij} = \begin{cases} \left(\ell_{\mathrm{R}}(i) - \ell^*_{\mathrm{R}}(i) \right) p_{R}(j|i), & \mathrm{for}\; i \in \mathcal{P}, j \in \mathcal{N} \\
    \left(\ell_{\mathrm{S}}(i) - \ell^*_{\mathrm{S}}(i) \right) p_{S}(j|i), & \mathrm{for}\;i \in \mathcal{P}, j \in \mathcal{P},\\
    0, & \mathrm{otherwise},
    \end{cases}
\end{align}
where ranking and sorting errors on $i$ are uniformly distributed with ranking pmf $p_R(j|i)$ and sorting pmf $p_S(j|i)$: 
\begin{align}\footnotesize
    \label{eq:RSpmf}
        p_{R}(j|i) = \frac{\mathrm{H}(x_{ij})}{\sum \limits_{k \in \mathcal{N}}\mathrm{H}(x_{ik})},
\end{align}
\begin{align}
    p_{S}(j|i) = \frac{\mathrm{H}(x_{ij})[IoU_j < IoU_i]}{\sum \limits_{k \in \mathcal{P}} \mathrm{H}(x_{ik})[IoU_k < IoU_i]}.
\end{align}
Similar to what we did for AP Loss, replacing $L_{ij}$ in Eq. 4, we obtain the gradient of RS Loss \cite{RSLoss} for $i$ $\in$ $\mathcal{P}$ as follows:
\begin{equation}
    \label{eq:GradientsRSinP} \footnotesize
        \frac{\partial\mathcal{L}_{\mathrm{RS}}}{\partial s_i} = \frac{1}{|\mathcal{P}|} \Big( \underbrace{\ell^*_{\mathrm{RS}}(i)-\ell_{\mathrm{RS}}(i)}_{\text{Update signal to promote $i$}} +\underbrace{ \sum \limits_{j \in \mathcal{P} } \left( \ell_{\mathrm{S}}(j)- \ell^*_{\mathrm{S}}(j) \right)  p_{S}(i|j)} _{\text{Update signal to demote $i$}} \Big).
\end{equation}
Doing the same for $i$ $\in$ $\mathcal{N}$ yields:
    \begin{align}
    \label{eq:RSNegGradients}
       \frac{\partial \mathcal{L}_{\mathrm{RS}}}{\partial s_i} = \frac{1}{|\mathcal{P}|}\sum \limits_{j \in \mathcal{P}} \ell_{\mathrm{R}}(j)p_{R}(i|j),
    \end{align}
which completes the derivation of the gradients for RS Loss.

\section{Further Details and Theoretical Discussion of Bucketed Ranking Based Losses}
This section presents further details on our loss functions and the proofs of Theorems 1 and 2 in the main paper.

\subsection{The Gradients of Bucketed AP Loss}
While obtaining the gradients of Bucketed AP (BAP) Loss, we follow the Identity Update framework, which basically requires the definition of the primary terms.
We already defined the primary terms of BAP Loss in Eq. 12 as:
\begin{equation}\footnotesize
\label{eq:s_BRanking_Error_Update}
    L^{b}_{ij}= \frac{\sum_{j=1}^i \mathrm{H}(x^b_{ij}) b_j}{\sum_{j=1}^i \mathrm{H}(x_{ij}) + \mathrm{H}(x^b_{ij}) b_j} \times \frac{b_{j}}{{N}_\mathrm{FP}(i)} , \quad i \in \mathcal{P}, j \in \mathcal{\Tilde{N}}
\end{equation}
and $L^{b}_{ij}=0$ if $i \in \mathcal{P}, j \in \mathcal{\Tilde{N}}$ does not hold. 
Please note that $\mathcal{\Tilde{N}}$ refers to the set of prototype negatives, not the actual negatives. The primary terms can be more compactly stated as:

\begin{align}
    \label{eq:PrimaryTermBAP}
    L^{b}_{ij}= \begin{cases} \left( \ell^b_R(i)  p(j^b|i)\right), & \quad i \in \mathcal{P}, j \in \mathcal{\Tilde{N}}, \\
    0, & \mathrm{otherwise},
    \end{cases}
\end{align}
in which 
\begin{align}
\label{eq:ell_r_bucketed}
\ell^b(i) = \frac{\sum_{j=1}^i \mathrm{H}(x^b_{ij}) b_j}{\sum_{j=1}^i \mathrm{H}(x_{ij}) + \mathrm{H}(x^b_{ij}) b_j},
\end{align}
and $p(j^b|i) = \frac{b_{j}}{{N}_\mathrm{FP}(i)}$.

Following Identity Update, the gradients for the $i$th logit is defined as:
\begin{equation}\footnotesize
    \label{eq:s_Final_Gradients}
    \frac{\partial \mathcal{L}}{\partial s_i} 
    = \frac{1}{Z} \Big( \sum \limits_{j} L_{ji} - \sum \limits_{j} L_{ij} \Big). 
\end{equation}
Therefore, we need to replace $L_{ij}$ by the primary term of the BAP Loss and set $Z=|\mathcal{P}|$ to find the gradients.
First, we do this for the positive logit $s_i$ in the following.
Replacing $L_{ij}$ by  $L^b_{ij}$ and  for $i \in \mathcal{P}$ we obtain:
\begin{align}
     \frac{\partial\mathcal{L_{BAP}}}{\partial s_i} = \frac{1}{|\mathcal{P}|} \left( \sum_j L^b_{ji} - \sum_j L^b_{ij} \right)
\end{align}
\begin{equation}
\label{eq:SummationOfPrimaryBAPPos}
    =  \frac{1}{|\mathcal{P}|} \left(  \sum_{j \in \mathcal{P}} L^b_{ji} + \sum_{j \in \mathcal{\Tilde{N}}} L^b_{ji} - \left( \sum_{j \in \mathcal{P}} L^b_{ij} + \sum_{j \in \mathcal{\Tilde{N}}} L^b_{ij} \right)  \right)
\end{equation}
Since there is no error defined among positives in AP and BAP loss, $\sum_{j \in \mathcal{P}} L^b_{ji}$ and $\sum_{j \in \mathcal{P}} L^b_{ij} = 0$. $L^b_{ji}$ is also equal to zero from our primary term definition (\cref{eq:PrimaryTermBAP}). Hence, \cref{eq:SummationOfPrimaryBAPPos} becomes:  
\begin{align}
     \frac{1}{|\mathcal{P}|} \left(  \cancelto{0}{\sum_{j \in \mathcal{P}} L^b_{ji}} + \cancelto{0}{\sum_{j \in \mathcal{\Tilde{N}}} L^b_{ji}} - \left( \cancelto{0}{\sum_{j \in \mathcal{P}} L^b_{ij}} + \sum_{j \in \mathcal{\Tilde{N}}} L^b_{ij} \right)  \right),
\end{align}
\begin{align}
    = -  \frac{1}{|\mathcal{P}|} \sum_{j \in \mathcal{\Tilde{N}}} L^b_{ij} = -  \frac{1}{|\mathcal{P}|}  \sum_{j \in \mathcal{\Tilde{N}}} \ell^b_R(i)  p(j^b|i) ,  
\end{align}
\begin{align}
    = -  \frac{1}{|\mathcal{P}|}  \ell^b_R(i) \sum_{j \in \mathcal{\Tilde{N}}}  p(j^b|i).
\end{align}
Considering that $\sum_{j \in \mathcal{\Tilde{N}}}  p(j^b|i)=1$ as $p(j^b|i)$ is a probability mass function,
\begin{align}
\label{eq:BAPGradientPositive}
    \frac{\partial\mathcal{L_{BAP}}}{\partial s_i} = -  \frac{1}{|\mathcal{P}|}  \ell^b_R(i), \quad \text{ if }  i \in \mathcal{P}
\end{align}
Now we follow Identity Update considering that $s_i$ is a \textit{prototype-negative logit}, in which case we have:
%
%
\begin{align} \footnotesize
\label{eq:SummationOfPrimaryBAP}
    \frac{\partial\mathcal{L_{BAP}}}{\partial s_i} =  \frac{1}{|\mathcal{P}|} \left(  \sum_{j \in \mathcal{P}} L^b_{ji} + \cancelto{0}{\sum_{j \in \mathcal{\Tilde{N}}} L^b_{ji}} - \left( \cancelto{0}{\sum_{j \in \mathcal{P}} L^b_{ij}} + \cancelto{0}{\sum_{j \in \mathcal{\Tilde{N}}} L^b_{ij}} \right)  \right).
\end{align}
The three terms in the equation cancel to $0$, considering that the primary term $L_{ij}$ can be positive only in the case when $i$ is a positive and $j$ is a negative (more precisely, prototype negative in this case).
Then, the following expression yields the gradients for the prototype negative:
\begin{align} \footnotesize
    \frac{\partial\mathcal{L_{BAP}}}{\partial s_i}  =  \frac{1}{|\mathcal{P}|}   \sum_{j \in \mathcal{P}} L^b_{ji} = \frac{1}{|\mathcal{P}|}  \sum_{j \in \mathcal{P}} \ell^b_R(j)  p(i^b|j).
\end{align}
However, we need to obtain the gradient for $k$th negative. To this end, we simply distribute the gradient considering the size of the bucket containing the $k$th negative, that is the $i$th bucket.
Representing the bucket size by $b_i$, and with a uniform distribution from the $i$th prototype negative over the $k$th negative in the $i$th bucket, the gradient for the $k$th negative is then:
\begin{align}
    \label{eq:BAPGradientNegative}
    \frac{\partial\mathcal{L_{BAP}}}{\partial s_k} = \frac{1}{|\mathcal{P}|}  \sum_{j \in \mathcal{P}} \ell^b_R(j)  p(i^b|j)  \frac{1}{b_i},
\end{align}
concluding the derivation of the gradients for Bucketed AP Loss.




\subsection{The Gradients of Bucketed RS Loss} \label{subsec:GradBRS}
In this section, we describe how we apply our bucketing approach to RS Loss, first for the negative logits and then for the positive logits, similar to what we did for AP Loss in the previous section.
Comparing Eq. \ref{eq:RSNegGradients} and \ref{eq:APGradient}, one can easily note that the gradient of the RS Loss is equal to the gradient of AP Loss if $s_k$ is a negative logit. 
In this case, one can refer to our derivation for Bucketed AP Loss, as they are identical.
Consequently, for $i \in \mathcal{N}$, using the same notation from the previous section, the gradient of the Bucketed RS Loss is:
\begin{align}
        \frac{\partial\mathcal{L_{BRS}}}{\partial s_k} 
    =  \frac{1}{|\mathcal{P}|}  \sum_{j \in \mathcal{P}} \ell^b_R(j)  p(i^b|j)  \frac{1}{b_i}  . 
\end{align}

\begin{figure*} 
    \centerline{
        \includegraphics[width=\textwidth]{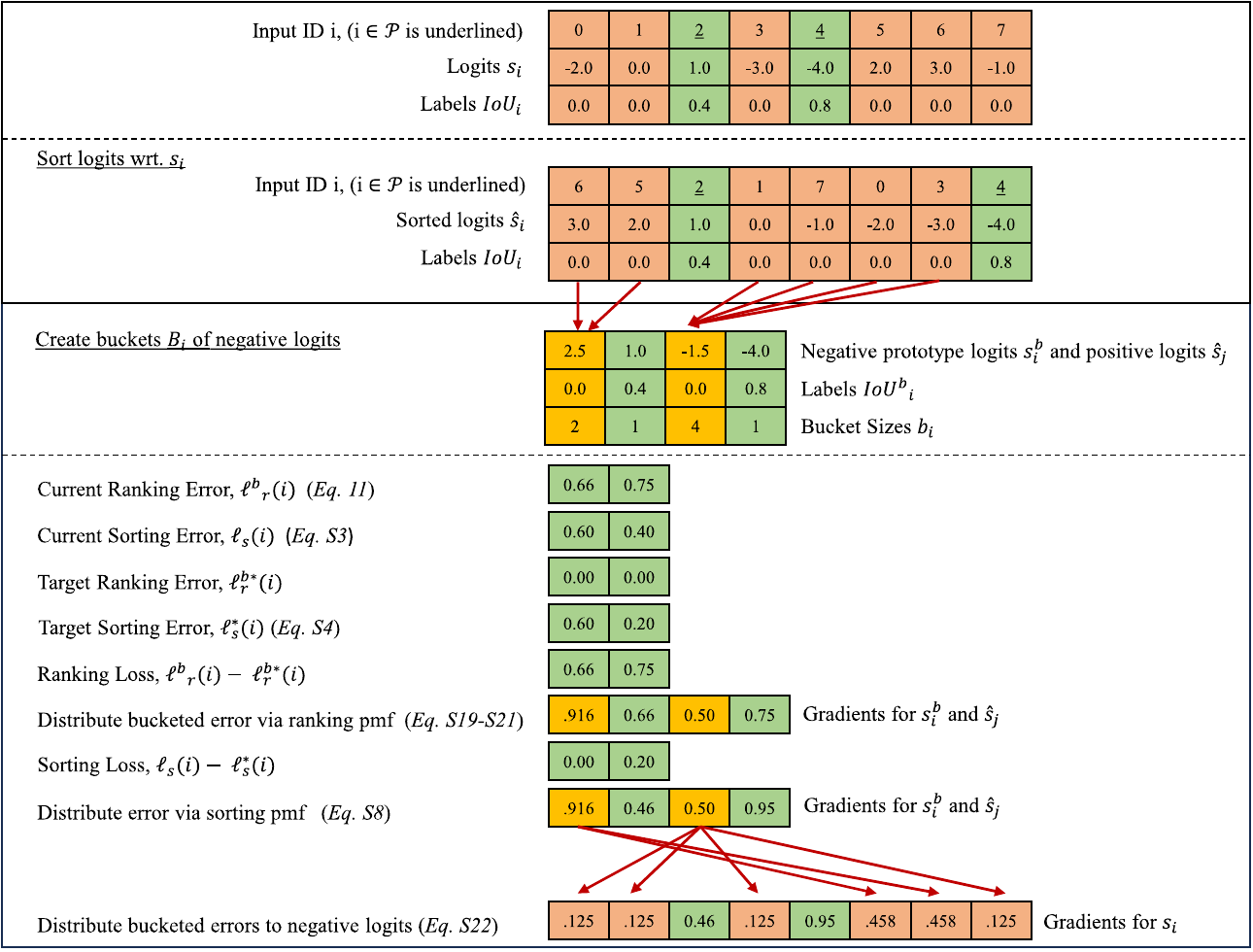}
    }
    \caption{An example illustrating the computation of gradients in BRS Loss. The colors green, red, and orange represent positive, negative, and prototype negative logits, respectively.
    \label{fig:BRSLoss_def}
} 
\end{figure*}

For $i \in \mathcal{P}$, the gradient of RS Loss is expressed in Eq. S8
\begin{align}
 \footnotesize
 &\frac{1}{|\mathcal{P}|} \Big( \ell^*_{\mathrm{RS}}(i)-\ell_{\mathrm{RS}}(i) + \sum \limits_{j \in \mathcal{P} } \left( \ell_{\mathrm{S}}(j)- \ell^*_{\mathrm{S}}(j) \right)  p_{S}(i|j) \Big),
\end{align}
which can be decomposed as:
\begin{align}
 = &\frac{1}{|\mathcal{P}|} \Big( \ell^*_{\mathrm{R}}(i)-\ell_{\mathrm{R}}(i) + \ell^*_{\mathrm{S}}(i)-\ell_{\mathrm{S}}(i) +  \\
 &\quad \sum \limits_{j \in \mathcal{P} } \left( \ell_{\mathrm{S}}(j)- \ell^*_{\mathrm{S}}(j) \right)  p_{S}(i|j) \Big), \\
 =&\frac{1}{|\mathcal{P}|} \Big( -\ell_{\mathrm{R}}(i) + \ell^*_{\mathrm{S}}(i)-\ell_{\mathrm{S}}(i) + \\
 &\quad \sum \limits_{j \in \mathcal{P} } \left( \ell_{\mathrm{S}}(j)- \ell^*_{\mathrm{S}}(j) \right)  p_{S}(i|j) \Big) .
\end{align}
Please note that calculating the sorting errors above ($\ell_{\mathrm{S}}(j)$, $\ell^*_{\mathrm{S}}(j)$) requires only positives, which can be efficiently computed. Then, the only term requiring the relations between the positives and the negatives is $\ell_{\mathrm{R}}(i)$.
We simply replace this term by the bucketed ranking error $\ell^b_{\mathrm{R}}(i)$ in Eq. \ref{eq:ell_r_bucketed}.
Consequently, the resulting gradient for the $i$th positive is:
\begin{align} \label{eq:BRSPosGrad}
\frac{1}{|\mathcal{P}|} \Big( -\ell^b_{\mathrm{R}}(i) + \ell^*_{\mathrm{S}}(i)-\ell_{\mathrm{S}}(i) + \sum \limits_{j \in \mathcal{P} } \left( \ell_{\mathrm{S}}(j)- \ell^*_{\mathrm{S}}(j) \right)  p_{S}(i|j) \Big).
\end{align}
%
This concludes the derivation of the gradients for Bucketed RS Loss. Please see Fig. \ref{fig:BRSLoss_def} for an example in which we illustrate obtaining the gradient of BRS Loss.

\subsection{Proof of Theorem 1}
\begin{theorem*}\label{s_theorem1}
Bucketed AP Loss and Bucketed RS Loss provide exactly the same gradients with AP and RS Losses respectively when $\delta=0$.
\begin{proof}
We divide the theorem into two parts, in which we first show that the gradients of Bucketed AP Loss reduce to those of AP Loss once $\delta=0$. And this will be followed by RS Loss.

\textit{Case 1. On the equality of the gradients for AP Loss and Bucketed AP Loss.} Similar to how we obtained the gradients of AP Loss, we investigate this for positive and negative logits respectively:

\underline{Case 1a. $s_i$ is a positive logit.} In this case, the gradient of Bucketed AP Loss is defined in Eq. \ref{eq:BAPGradientPositive} as follows:
\begin{align}
    \frac{\partial\mathcal{L_{BAP}}}{\partial s_i} = -  \frac{1}{|\mathcal{P}|}  \ell^b_R(i) \text{, \quad if }  i \in \mathcal{P},
\end{align}
where, by denoting the pair-wise relations between the $i$th positive and the $j$th prototype negative by $\mathrm{H}(x^b_{ij})$ and the size of the $j$th bucket $b_j$, the bucketed ranking error is:
\begin{align}
\ell_R^b(i) &= \frac{\sum_{j=1}^i \mathrm{H}(x^b_{ij}) b_j}{\sum_{j=1}^i \mathrm{H}(x_{ij}) + \mathrm{H}(x^b_{ij}) b_j}, \\
&= \frac{\sum_{j=1}^i \mathrm{H}(x^b_{ij}) b_j}{\sum_{j=1}^i \mathrm{H}(x_{ij}) + \sum_{j=1}^i \mathrm{H}(x^b_{ij}) b_j}, \\
\end{align}
Considering that $\mathrm{H}(x^b_{ij})=0$ for $j>i$, we can express the bucketed ranking error as follows:
\begin{align}
\label{eq:ell_ranking_proof}
\ell_R^b(i)= \frac{\sum_{j \in \Tilde{\mathcal{N}}} \mathrm{H}(x^b_{ij}) b_j}{\sum_{j \in  \mathcal{P}} \mathrm{H}(x_{ij}) + \sum_{j \in \Tilde{\mathcal{N}}} \mathrm{H}(x^b_{ij}) b_j}.
\end{align}%
Please note that in the case that $\delta=0$, the step function $\mathrm{H}(\cdot)$ corresponds to (Eq. 2 in the paper) the following form:
\begin{equation}\footnotesize
\label{eq:s_h_smooth}
H(x)=\left\{
\begin{aligned}
&0\, , & x < 0 \\
&1\, , & 0 < x,
\end{aligned}
\right.
\end{equation}
where we assume that no logits are exactly equal (i.e., $x\neq 0$) for the sake of simplicity. As a result, the term $\sum_{j \in \Tilde{\mathcal{N}}} \mathrm{H}(x^b_{ij}) b_j$ both in the numerator and the denominator of Eq. \ref{eq:ell_ranking_proof} can be expressed as:
\begin{align}
    \sum_{j \in \Tilde{\mathcal{N}}} \mathrm{H}(x^b_{ij}) b_j = \sum_{j\in \mathcal{N}} H(x_{ij}).
\end{align}
This is because, for any prototype logit $s^b_j$,  $s_{j-1} > s^b_j >s_{j}$ holds and the cardinality of the $j$th bucket is $b_j$. In other words, counting the number of negatives one-by-one as in the right-hand side of the equation versus counting each bucket size of negatives and summing over the bucket sizes (in the left-hand side expression) are equal. Furthermore, these expressions both yield the number of false positives on the $i$th positive examples, i.e., ${N}_\mathrm{FP}(i)$.
Consequently, replacing this term in Eq. \ref{eq:ell_ranking_proof} shows that the bucketed ranking error is equal to the ranking error when $\delta=0$:
\begin{align}
\ell^b_R(i)&= \frac{\sum_{j \in \Tilde{\mathcal{N}}} \mathrm{H}(x^b_{ij}) b_j}{\sum_{j \in  \mathcal{P}} \mathrm{H}(x_{ij}) + \sum_{j \in \Tilde{\mathcal{N}}} \mathrm{H}(x^b_{ij}) b_j}, \\
&= \frac{\sum_{j\in \mathcal{N}} H(x_{ij})}{\sum_{j \in  \mathcal{P}} \mathrm{H}(x_{ij}) + \sum_{j\in \mathcal{N}} H(x_{ij})}, \\
&= \ell_R(i).
\end{align}
As a result:
\begin{equation}
    \frac{\partial\mathcal{L_{BAP}}}{\partial s_i} = -  \frac{1}{|\mathcal{P}|}  \ell^b_R(i)= -  \frac{1}{|\mathcal{P}|}  \ell_R(i)=\frac{\partial\mathcal{L_{AP}}}{\partial s_i} \text{, \quad if }  i \in \mathcal{P},
\end{equation}

\underline{Case 1b. $s_i$ is a negative logit.} For this case, the gradient expression of the Bucketed AP Loss is presented in Eq. \ref{eq:BAPGradientNegative} as follows:
\begin{align}
    \label{eq:BAPGradientNegative_proof}
    \frac{\partial\mathcal{L_{BAP}}}{\partial s_i} = \frac{1}{|\mathcal{P}|}  \sum_{j \in \mathcal{P}} \ell^b_R(j) p(k^b|j)  \frac{1}{b_k},
\end{align}
where $k$ represents the bucket containing the $i$th negative. In case 1a, we have already shown that $\ell^b_R(j)=\ell_R(j)$ when $\delta=0$.
Hence, now we focus on the $p(i^b|j)  \frac{1}{b_i}$, which reduces to:
\begin{align}
p(k^b|j)  \frac{1}{b_k}=\frac{b_{k} }{ {N}_\mathrm{FP}(j)}  \frac{1}{b_i}=\frac{1 }{ {N}_\mathrm{FP}(j)}.
\end{align}
Please note that, for a negative logit $s_i$ with a higher value than $s_j$, which is the case for the probability mass function of AP Loss, $1=\mathrm{H}(x_{ij})$ holds.
As a result:
\begin{align}
 \frac{1 }{ {N}_\mathrm{FP}(j)} =  \frac{\mathrm{H}(x_{ij})}{ {N}_\mathrm{FP}(j)} = p(i|j).
\end{align}
Finally, replacing these terms in Eq. \ref{eq:BAPGradientNegative_proof}, we have
\begin{align}
    \frac{\partial\mathcal{L_{BAP}}}{\partial s_i} &= \frac{1}{|\mathcal{P}|}  \sum_{j \in \mathcal{P}} \ell^b_R(j)  p(k^b|j)  \frac{1}{b_k} \\
    &=\frac{1}{|\mathcal{P}|}  \sum_{j \in \mathcal{P}} \ell_R(j)  p(i|j) = \frac{\partial\mathcal{L_{AP}}}{\partial s_i},
\end{align}
completing the proof for Bucketed AP Loss.

\textit{Case 2. On the equality of the gradients for RS Loss and Bucketed RS Loss.} Similar to how we obtained the gradients of AP Loss, we investigate this for the positive and negative logits respectively:

\underline{Case 2a. $s_i$ is a positive logit.} Please note that the gradient of RS Loss in this case is defined in Eq. \ref{eq:BRSPosGrad}. As we have already shown in Case 1a that $\ell^b_R(j)=\ell_R(j)$ when $\delta=0$ and we do not modify the remaining terms, this case holds.

\underline{Case 2b. $s_i$ is a negative logit.} Similarly, when the logit is negative, the gradient of RS Loss is equal to the gradient of AP Loss as we discussed in Section \ref{subsec:GradBRS}. Following from Case 1b, this case also holds, completing proof of the theorem.

\end{proof}
\end{theorem*}
\subsection{Proof of Theorem 2}
\begin{theorem*}
\label{s_s_s_theorem2}
Bucketed RS and Bucketed AP Losses have $\mathcal{O}(\max ((|\mathcal{P}|+|\mathcal{N}|) \log(|\mathcal{P}|+|\mathcal{N}|), |\mathcal{P}|^2))$ time complexity.

\newcommand{\BigO}[1]{\(\mathcal{O}(#1)\)}

\begin{proof}
We first summarize the time complexity for each line of \cref{alg:bucketed_RS} and then provide the derivation details.

\begin{center}\small 
\begin{tabular}{cc} \hline
   Line in Alg. \ref{alg:bucketed_RS} & Time Complexity \\ \hline\hline
   Line 1  & \BigO{(|\mathcal{P}|+|\mathcal{N}|) \log(|\mathcal{P}|+|\mathcal{N}|)} \\ 
   Line 2  & \BigO{|\mathcal{P}|+|\mathcal{N}|} \\ 
   Line 3  & \BigO{|\mathcal{P}|^2} \\ 
   Line 4  & \BigO{|\mathcal{P}|^2} \\ 
   Line 5  & \BigO{|\mathcal{P}|^2} \\ 
   Line 6  & \BigO{|\mathcal{P}|^2} \\ 
   Line 7  & \BigO{|\mathcal{P}|^2} \\ 
   Line 8  & \BigO{|\mathcal{P}|^2} \\ 
   Line 9  & \BigO{|\mathcal{P}|^2} \\ 
   Line 10  & \BigO{|\mathcal{P}|^2} \\ 
   Line 11  & \BigO{|\mathcal{P}|+|\mathcal{N}|} \\ 
   Line 12  & \BigO{|\mathcal{P}|+|\mathcal{N}|} \\  \hline
   Overall & {\small \BigO{\max( (|\mathcal{P}|+|\mathcal{N}|) \log(|\mathcal{P}|+|\mathcal{N}|), |\mathcal{P}|^2 ) } } \\ \hline  
\end{tabular}
\end{center}

Now we provide the derivation details for each line: 

\paragraph{\cref{al2:sort-logits}: Sorting  logits.} For sorting the logits, we use \textbf{torch.sort}, which internally employs the Quicksort implementation of NumPy (\textbf{np.sort}). Therefore, the time complexity of \cref{al2:sort-logits} is \(\mathcal{O}\left( |\mathcal{S}| \log |\mathcal{S}| \right)\), where \(|\mathcal{S}| = |\mathcal{P} \cup \mathcal{N}| = |\mathcal{P}| + |\mathcal{N}| \) is the length of the array to be sorted. Since $|\mathcal{P}|$ is negligible ($|\mathcal{P}| \ll |\mathcal{N}|$), the time complexity at \cref{al2:sort-logits} can be approximated as \(\mathcal{O}\left( |\mathcal{N}| \log |\mathcal{N}| \right)\).

\paragraph{\cref{al2:bucket-sorted-logits}: Bucketing.} Bucketing the sorted array of \(\hat{s}_1\), \(\hat{s}_2\), $....$, \(\hat{s}_{|\mathcal{S}|}\) can be computed in one pass. Therefore, its time complexity is \BigO{|\mathcal{S}|} $\approx$ \BigO{|\mathcal{N}|} since the length of the pass is $\mathcal{S}| = |\mathcal{P} \cup \mathcal{N}| = |\mathcal{P}| + |\mathcal{N}|$ and $|\mathcal{P}| \ll |\mathcal{N}|$.

\paragraph{Line \ref{al2:difference-terms}: Difference Terms.}
For one $i \in \mathcal{P}$ and $\forall j^b \in \Tilde{\mathcal{N}}$, the time complexity of the term  $x_{ij}^{b} = s_{j}^{b} - s_i$ is constant, i.e. \BigO{1}. Therefore, the worst-case time complexity of Line \ref{al2:difference-terms} is \BigO{|\mathcal{P}| \cdot |\Tilde{\mathcal{N}}|} $\approx$ \BigO{|\mathcal{P}|^2)} since \(|\Tilde{\mathcal{N}}| \le |\mathcal{P}| + 1\).

\paragraph{Line \ref{al2:current-ranking-error}: Current Bucketed Ranking Error.} 
The worst-case scenario for the current ranking error $\ell^{b}_{\mathrm{R}}$ is when $i=|\Tilde{\mathcal{S}}| = 2|\mathcal{P}| + 1$ since the worst-case scenarios for both $\mathrm{N_{FP}}(i) = \sum_{j = 1}^{i} \mathrm{H}(x_{ij}^{b})b_{j}$ and $\mathrm{rank}(i) = \sum_{j = 1}^{i} \mathrm{H}(x_{ij}) + \mathrm{H}(x_{ij}^{b})b_{j}$ are when $i=|\Tilde{\mathcal{S}}| = 2|\mathcal{P}| + 1$. That is, the time complexity of $\mathrm{N_{FP}}(|\Tilde{\mathcal{S}}|)$ and $\mathrm{rank}(|\Tilde{\mathcal{S}}|)$ is \BigO{|\Tilde{\mathcal{S}}|} = \BigO{|\mathcal{P}|}, so is the time complexity of the current ranking error. Therefore, computing the current ranking loss for $\mathcal{|P|}$ positive logits has a time complexity of \BigO{|\mathcal{P}|^2}.

\paragraph{Line \ref{al2:bucketed-primary-terms}: Bucketed Primary Terms.} 
The target ranking error $\ell^{*b}_{\mathrm{R}} = 0$, making its complexity constant, i.e. \BigO{1}.
Therefore, the computation of the primary terms $L^{b}_{ij}$ = $\left(\ell_{\mathrm{R}}(i) - \ell^*_{\mathrm{R}}(i) \right) p_{R}(j|i)$ for every $i \in \mathcal{P}$ and every $j \in \Tilde{\mathcal{N}}$ has a worst-case time complexity of \BigO{|\mathcal{P}| \cdot |\tilde{N}|}. Since \(|\Tilde{\mathcal{N}}| \le |\mathcal{P}| + 1\), the complexity of Line \ref{al2:bucketed-primary-terms} can be approximated as \BigO{|\mathcal{P}|^2}.

\paragraph{Line \ref{al2:current-sorting-error} and Line \ref{al2:target-sorting-error}: Bucketed Sorting Error.}
The time complexity of the current sorting error $\ell^{b}_{\mathrm{S}}$ at the worst case is the maximum of time complexities of the numerator and denominator terms in Eq. \ref{eq:RSCurrent}. 
The numerator is the summation of terms with constant time complexities. The denominator, $\mathrm{rank}^+(i)$, has a time complexity of \BigO{|\mathcal{P}|} in the worst case, similar to $\mathrm{rank}(i)$. 
Therefore, the worst-case time complexity of the current sorting error is \BigO{|\mathcal{P}|}.
Similar to the current sorting error, both the numerator and the denominator of Eq. \ref{eq:RSTarget} are summations of the terms with constant time complexities. Therefore, the worst-case time complexity of the target sorting error is \BigO{|\mathcal{P}|} as well.
Lastly, computing the current sorting loss $\ell^{b}_{\mathrm{S}}$ and the target sorting loss $\ell^{*b}_{\mathrm{S}}$ for every positive logits corresponds to \BigO{|\mathcal{P}|^2}.

\paragraph{Line \ref{al2:primary-terms}: Primary Terms.}
Similar to Line \ref{al2:bucketed-primary-terms}, the computation of the primary terms $L_{ij}$ = $\left(\ell^{b}_{\mathrm{S}}(i) - \ell^{*b}_{\mathrm{S}}(i) \right) p_{S}(j|i)$ for every $i, j \in \mathcal{P}$ has a worst-case time complexity of \BigO{|\mathcal{P}| \cdot |\mathcal{P}|} =  \(O( |\mathcal{P}|^2 )\).

\paragraph{Line \ref{al2:final-gradients-for-positive}: Final Gradients for Positive Logits.}
Calculating the final gradients $\frac{\partial \mathcal{L}}{\partial s_i} = \frac{1}{Z} \sum \limits_{j \in \Tilde{\mathcal{S}}} \big( L_{ji} - L_{ij} \big)$ in Eq. \ref{eq:s_Final_Gradients} has a time complexity of \BigO{|\Tilde{\mathcal{S}}|} = \BigO{|\mathcal{P}|} since the primary terms are already calculated in Line \ref{al2:primary-terms} and \(|\Tilde{\mathcal{S}}| \le 2|\mathcal{P}| + 1\). Therefore, computing the final gradients for every positive logits is \BigO{ |\mathcal{P}|^2}.

\paragraph{Line \ref{al2:final-gradients-for-prototype}: Final Gradients for Negative Prototypes.}
Similar to Line \ref{al2:final-gradients-for-positive}, the time complexity of computing final gradients $\frac{\partial \mathcal{L}}{\partial s_i}$ for every negative prototype logits is \BigO{ |\Tilde{\mathcal{N}}|\cdot |\mathcal{P}|} = \BigO{|\mathcal{P}|^2} since $|\Tilde{\mathcal{N}}| \le |\mathcal{P}| + 1$.

\paragraph{Line \ref{al2:final-gradients-for-negative}: Final Gradients for Negative Logits.}
Distributing the final gradients computed for every negative prototype logits, i.e. $\Tilde{\mathcal{N}}$, to negative logits, i.e. $\mathcal{N}$, has a time complexity of \BigO{|\mathcal{N}|}, since it can be done in one pass over the sorted array \(\hat{s}_1\), \(\hat{s}_2\), $....$, \(\hat{s}_{|\mathcal{S}|}\) and related gradients for negative prototype logits are computed already in Line 11.

\paragraph{Line \ref{al2:normalize}: Normalization.}
Applying normalization operation, with time complexity \BigO{1}, for each gradient corresponding to the logits in the sorted array \(\hat{s}_1\), \(\hat{s}_2\), $....$, \(\hat{s}_{|\mathcal{S}|}\) has a time complexity of \BigO{|\mathcal{S}|} = \BigO{|\mathcal{N}|} since $|\mathcal{S}| =|\mathcal{P} \cup \mathcal{N}| = |\mathcal{P}| + |\mathcal{N}|$ and $|\mathcal{P}| \ll |\mathcal{N}|$. 

\paragraph{Overall time complexity of Algorithm \ref{alg:bucketed_RS}.}
The time complexity of the whole algorithm is the sum of the time complexity at each line, which equals to the time complexity of the line with the maximum time complexity. That is, \BigO{\max( (|\mathcal{P}|+|\mathcal{N}|) \log( |\mathcal{P}|+|\mathcal{N}|), |\mathcal{N}|, |\mathcal{P}|^2)} = \BigO{\max( (|\mathcal{P}|+|\mathcal{N}|) \log( |\mathcal{P}|+|\mathcal{N}|),  |\mathcal{P}|^2)}.
\end{proof}
\end{theorem*}

\section{Implementation Details}
This section includes the implementation details of our approach.

\subsection{Implementation Details for Multi-stage and One-stage Detectors}

In Section 6.1.1 of the paper for both COCO \cite{COCO} and LVIS \cite{LVIS}, we strictly follow previous work to provide a fair comparison. Specifically, similar to RS Loss \cite{RSLoss}, while training multi-stage detection and segmentation methods, i.e., Faster R-CNN \cite{FasterRCNN}, Cascade R-CNN \cite{CascadeRCNN} and Mask R-CNN \cite{MaskRCNN}, we do not use random sampling different from the original architecture. We use all anchors for RPN and top-1000 proposals for Faster R-CNN \cite{FasterRCNN} and top-2000 proposals for Cascade R-CNN \cite{CascadeRCNN}. Again similar to RS Loss, we replace softmax classifier with class-wise binary sigmoid classifiers. The initial learning rate is set to 0.012 and decreased by a factor of 10 after the 8th and 11th epochs for the experiments we train the models for 12 epochs. Again, following RS Loss and AP Loss, we follow the same scheduling but with an initial learning rate of 0.008 respectively for the experiments with one-stage object detectors.

For experiments with Cityscapes \cite{Cordts2016Cityscapes}, we follow the Mask R-CNN\cite{MaskRCNN} configuration provided in mmdetection and replace the loss function with Bucketed RS Loss and RS Loss. 



\subsection{Further Details for the Analyses with Synthetic Data}
In our analyses using the synthetic data in Section 6.1.2,  we generate logits with different cardinalities $L=\{10K, 100K, 1M\}$ also for various percentages of positives $m=\{0.1, 1.0, 2.0, 5.0\}$. Specifically, we sample positive and negative logits from the Gaussian distributions, $s_i^+ \sim \mathcal{N}(-1, 1)$ and $s_i^- \sim \mathcal{N}(1, 1)$ respectively. Please note that the mean of the negative logits is higher than that of the positive logits. This ensures that the sampled set of logits is likely to include trivial cases in which all the positives have higher confidence than all the negatives, and consequently, the set  of negative logits (e.g., $\hat{\mathcal{N}}$ line 2 in Alg. 2) is empty and the loss computation is trivial. For RS and Bucketed RS-Loss experiments, we generated uniformly distributed random IoU values for positive logits. To illustrate the extreme case, one might consider $L=1M$ logits with $m=0.1$ correspond to anchor-based detector making dense predictions such as RetinaNet \cite{FocalLoss}, the smaller number of logits mimics the R-CNN head of two-stage detectors, and finally  $L=10K$ with higher $m$ approximates recent transformer-based detectors.
For the sake of simplicity, here we compare our BAP Loss with AP Loss in terms of time efficiency and number of floating point operations (FLOPs), report the mean iteration time of three independent runs with each setting.

\begin{figure}
    \includegraphics[width=\textwidth]{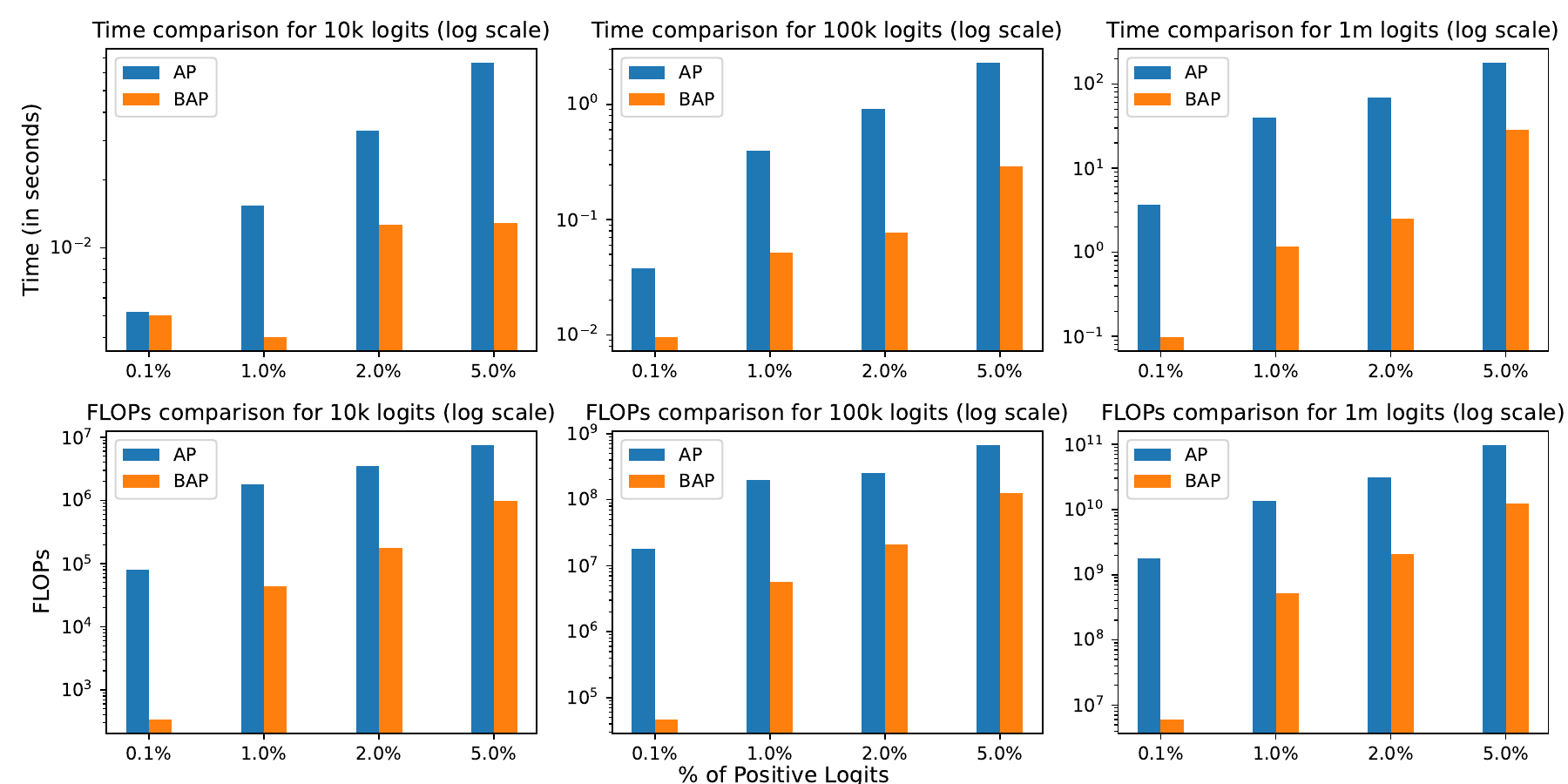}

    \vspace{-2.ex}
    \caption{Performance Comparison of AP and BAP Loss Functions: Log-Scale Analysis of Computational Time and number of floating point operations (FLOPs) across various Data cardinalities and percentages.}
    \label{fig:ap_bap_time}
 \end{figure}

Fig. \ref{fig:ap_bap_time} shows that our bucketing approach, in fact, improve the loss efficiency by up to $\sim 40\times$, a very significant improvement. Furthermore, when we focus on FLOPs of AP Loss decrease from 76M to 900k with our BAP Loss, introducing more than $80\times$ less FLOPs  and further validating the effectiveness of our approach. 

\subsection{Implementation Details for BRS-DETR}
We incorporate our BRS Loss into the official Co-DETR repository to be aligned with the original implementation and keep its settings unless otherwise explicitly stated. 
We train our BRS-DETR for 12 epochs on 8 Tesla A100 GPUs, with each GPU processing 2 images, resulting in a total batch size of 16. 
We initialize the learning rate to $0.0002$ and decrease it by a factor of $5$ after epochs 10 and 11.
We employ the setting of Co-DETR with 300 queries.
As for the auxiliary heads, we similarly use ATSS \cite{ATSS} and Faster-RCNN \cite{FasterRCNN}.
Different from the official implementation, we remove sampling in Faster R-CNN since BRS Loss is already robust to class imbalance among positive and negative proposals.
Similarly for ATSS, we do not use centerness head following RS-ATSS.
Following RS R-CNN and RS-ATSS \cite{RSLoss}, we use self-balancing and do not tune the weights of the localisation losses.
As described in Section 5, we also employ self-balancing in Co-DETR localisation losses.
Finally, once our loss function is used, we find it useful (i) to increase the classification loss weight used for Hungarian assignment of the proposals as positives or negatives to 4; and (ii) decrease the auxiliary head weight to 5.


%
%



\end{document}